\documentclass[]{fairmeta}

\title{Brittlebench: Quantifying LLM Robustness via Prompt Sensitivity}

\author[1,2,*]{Angelika Romanou}
\author[1]{Mark Ibrahim}
\author[1]{Candace Ross}
\author[1]{Kerem Oktar}
\author[1,3,*]{Chantal Shaib}
\author[1]{Kerem Oktar}
\author[1]{Samuel J. Bell}
\author[1]{Anaelia Ovalle}
\author[2]{Antoine Bosselut}
\author[1]{Jesse Dodge}
\author[1]{Koustuv Sinha}
\author[1]{Adina Williams}

\affiliation[1]{FAIR at Meta}
\affiliation[2]{EPFL}
\affiliation[3]{Northeastern University}

\contribution[*]{Work done at Meta}

\abstract{
Existing evaluation methods largely rely on clean, static benchmarks, which can overestimate true model performance by failing to capture the noise and variability inherent in real-world user inputs.
This is especially true for language models, which can face human-generated text queries containing mistakes, typos, or alternative ways of phrasing the same question.  
In this work, we introduce a theoretical framework for quantifying model sensitivity to prompt variants, or \emph{brittleness}, that can enable us to disentangle data-induced difficulty from prompt-related variability.
Using this framework, we design a novel evaluation pipeline, \bb{}, to holistically evaluate the sensitivity of frontier models. 
We apply semantics-preserving perturbations to a suite of popular benchmarks, and observe model average performance to degrade as much as $\sim$$21\%$.
However, these perturbations do not affect all models equally: even a single perturbation alters the relative ranking of models in 63\% of cases, impacting conclusions about comparative model performance.
Decomposing the total variance of both state-of-the-art open-weight and commercial models, we find that semantics-preserving input perturbations can account for up to half of the performance variance for a given model.
\bb{} highlights the need for more robust evaluations and models, and allows us to systematically understand model brittleness.}

\date{\today}
\correspondence{\email{romanou@meta.com} or \email{adinawilliams@meta.com}}

\metadata[Code]{\url{https://github.com/facebookresearch/Brittlebench}}

\begin{document}

\newcommand{\bb}{\texttt{Brittlebench}}

\definecolor{saffron}{RGB}{227, 170, 0}
\definecolor{pink}{RGB}{204, 0, 102}
\newcommand{\Candace}[1]{\textcolor{saffron}{\textit{(Candace: #1)}}}
\newcommand{\jdcomment}[1]{\textcolor{teal}{\textit{(Jesse: #1)}}}
\newcommand{\ks}[1]{\textcolor{blue}{\textit{(Koustuv: #1)}}}
\newcommand{\Mark}[1]{\textcolor{purple}{\textit{(Mark: #1)}}}
\newcommand{\AR}[1]{\textcolor{pink}{\textit{(Angie: #1)}}}

\definecolor{blue_back}{RGB}{207,226,243}
\definecolor{blue_agg}{RGB}{66,133,244}
\definecolor{green_back}{RGB}{217,234,211}
\definecolor{green_agg}{RGB}{106,168,79}
\definecolor{yellow_back}{RGB}{255,242,204}
\definecolor{orange_back}{RGB}{252,229,205}
\definecolor{purple_back}{RGB}{217,210,233}

\maketitle

\section{Introduction}
\label{section:intro}

Over recent years, the evaluation of AI systems has increasingly focused on fine-grained numerical comparisons, often down to marginal decimal differences, in benchmark scores. Across both research and production settings, model comparisons are often made over a small set of widely adopted benchmarks, such as MMLU \citep{hendryckstest2021}, ARC \citep{allenai:arc}, and TruthfulQA \citep{lin2021truthfulqa}, with each new model aiming to surpass its predecessors in the existing state of the art. 
Yet the same model evaluated on the same benchmark can swing several points from changes that leave the underlying task unchanged: a typo, a reformatted option list, an extra space between tokens \citep{sclar2024quantifying, mizrahi-etal-2024-state, voronov-etal-2024-mind, haller2025llm, qian-etal-2024-varbench}. 
Such sensitivity raises the fundamental question of whether benchmark gains reflect genuine improvements in capability, or increasing adaptation to benchmark-specific formatting artifacts \citep{gururangan-etal-2018-annotation, geirhos-etal-2020-shortcut}.

A growing body of work has documented this fragility across many axes of variation, including template-level changes such as paraphrased instructions or reordered few-shot examples \citep{sclar2024quantifying, mizrahi-etal-2024-state, voronov-etal-2024-mind}, character- and token-level perturbations such as typos and whitespace insertion \citep{moradi-samwald-2021-evaluating, sun2024evaluating, haller2025llm, salinas2024butterfly, prabhakaran-etal-2019-perturbation}, and structural choices such as multiple-choice option ordering \citep{gupta-etal-2024-changinganswerorderdecrease, wu2025rewordbench, habba2025dove}, as summarized in Table~\ref{tab:related_work}. 
A common response to these findings has been to assume that scale resolves them and that frontier models, trained on broader data and with more compute, will exhibit the stable, format-invariant behavior expected of genuinely capable systems. 
This assumption remains weakly tested, as existing perturbation studies focus predominantly on smaller open-weight models, and the few that include frontier systems do so over narrow perturbation sets or single benchmarks, leaving the scaling hypothesis empirically underdetermined.
Our results challenge this scaling hypothesis, showing that frontier models exhibit prompt sensitivity qualitatively similar to that of smaller open-weight models.

Despite extensive evidence of prompt sensitivity, no principled framework exists for quantifying how much observed performance variability arises from prompt formulation \citep{wu2025rewordbench}  rather than underlying task difficulty.
Moreover, benchmark creators often fail to assess whether their benchmark's construct validity might be compromised by arbitrary format effects or prompt-related biases \citep{bean2025measuring}. 
As a result, current evaluation practice lacks a principled way to determine whether benchmark scores reflect stable, underlying abilities or merely artifacts of prompt design. 

\newcommand{\cmk}{\includegraphics[height=1em]{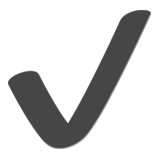}}

\begin{table*}
\centering
\resizebox{\textwidth}{!}{%
\begin{tabular}{l|ccccc|c|cc|c}
 & \multicolumn{5}{c|}{\textbf{SCOPE COVERAGE}}
 & \textbf{METHOD}
 & \multicolumn{2}{c|}{\textbf{ANALYSIS DEPTH}}
 & \textbf{ARTIFACTS} \\
\cmidrule(lr){2-10}
\textbf{Paper}
 & \rotatebox{25}{\colorbox{blue_back}{Word-level}}
 & \rotatebox{25}{\colorbox{purple_back}{Prompt-level}}
 & \rotatebox{25}{\colorbox{green_back}{Instructions-level}}
 & \rotatebox{25}{\colorbox{yellow_back}{Paraphrasing}}
 & \rotatebox{25}{\colorbox{orange_back}{Task-specific}}
 & \textbf{\begin{tabular}[c]{@{}c@{}}Variance\\decomposition\end{tabular}} 
 & \textbf{\begin{tabular}[c]{@{}c@{}}Reasoning\\analysis\end{tabular}} 
 & \textbf{\begin{tabular}[c]{@{}c@{}}Ranking \\stability\end{tabular}} 
 & \textbf{Software / Datasets}  \\[1.5ex]
\midrule
\texttt{DOVE} \citep{habba2025dove}              & -    & \cmk & \cmk & -    & -    & -      & -      & -      & \includegraphics[height=1em]{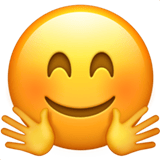} Dataset \\
\texttt{PromptBench} \citep{zhu2024promptbench}  & -    & -    & \cmk & -    & -  & -      & -    & -    & \includegraphics[height=1em]{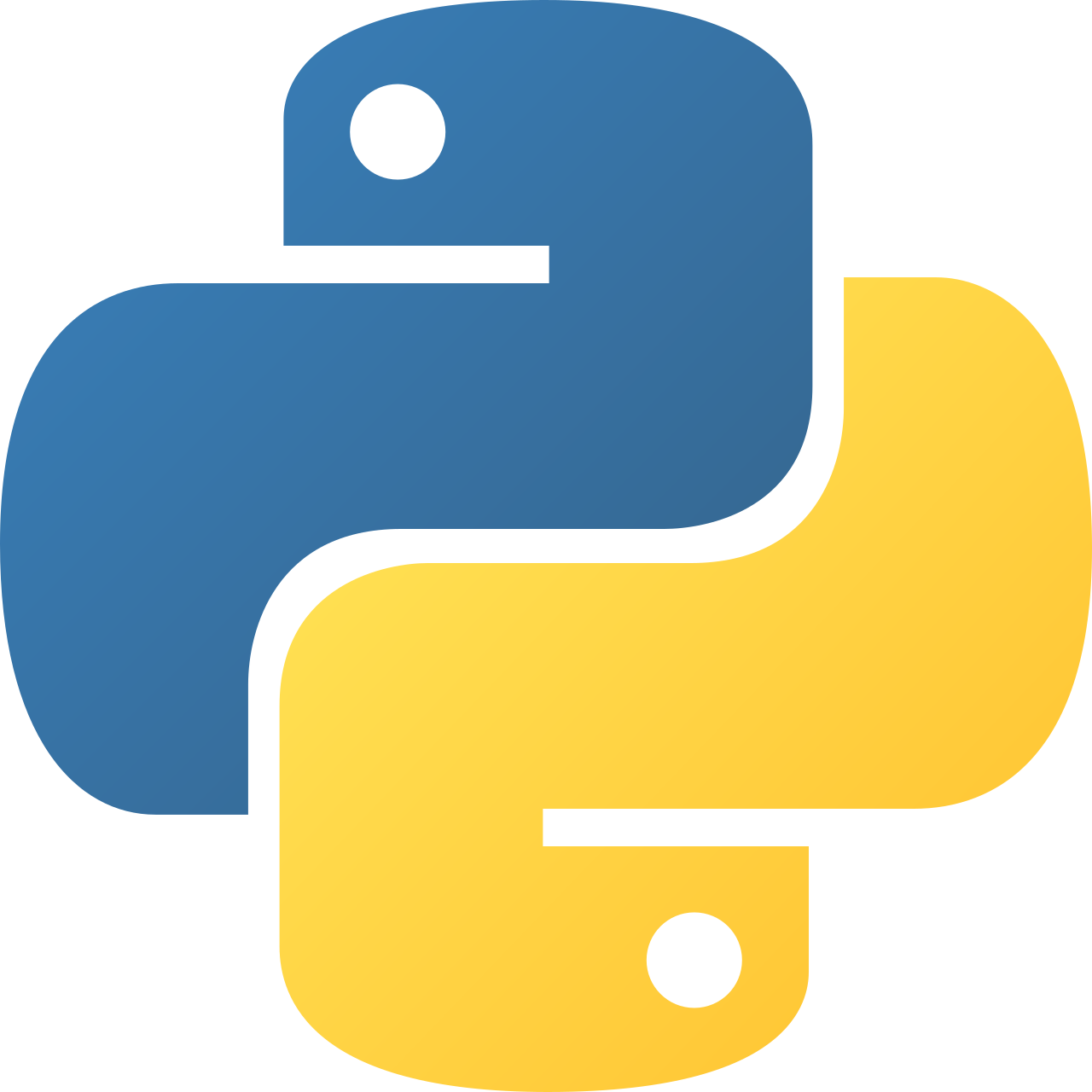} Python lib \\
\cite{mizrahi-etal-2024-state}                   & -    & -    & \cmk & -    & -    & -      & -      & \cmk & \includegraphics[height=1em]{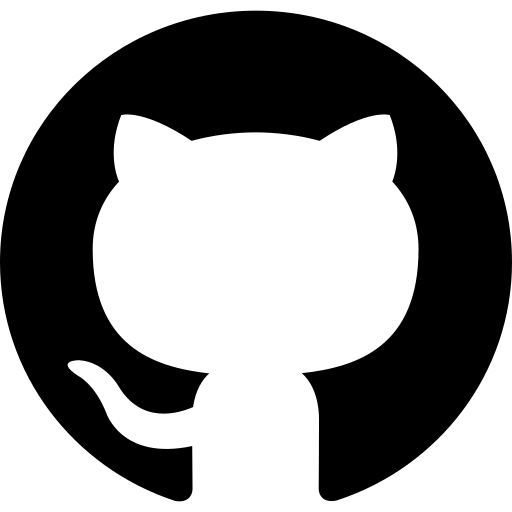} Code + Data \\
\cite{alzahrani2024benchmarks}                   & -    & \cmk & \cmk & -     & -      & \begin{tabular}[c]{@{}c@{}}Token bias vs.\\Position bias\end{tabular} & -    & \cmk & - \\
\cite{salinas2024butterfly}                      & \cmk & \cmk & -    & -     & -      & -      & -    & -    & - \\
\cite{zheng2023large}                            & -    & \cmk & -    & -    & -      & \begin{tabular}[c]{@{}c@{}}Token bias vs.\\Position bias\end{tabular} & \cmk & \cmk & \includegraphics[height=1em]{files/github.png} Code + Data \\
\cite{cao2024worst}                              & -    & -    & \cmk & -    & -      & Across 4 factors & - & \cmk & \includegraphics[height=1em]{files/github.png} Code + Data \\
\texttt{POSIX} \citep{chatterjee-etal-2024-posix}       & \cmk & -    & \cmk & \cmk & -        & -      & -    & -    & - \\
\texttt{ProSA} \citep{zhuo-etal-2024-prosa}      & -    & -    & \cmk & -      & -      & -      & -    & -    & - \\
\cite{voronov-etal-2024-mind}                    & \cmk & \cmk & -    & -     & -      & -      & -    & \cmk & - \\
\texttt{TextFlint} \citep{wang2021textflint}     & \cmk & -    & -    & \cmk & \cmk & -      & -       & -    & \includegraphics[height=1em]{files/github.png} Code + Data \\
\midrule
\textbf{\bb{} (OURS)}
 & \cmk
 & \cmk
 & \cmk
 & \cmk
 & \cmk
 & \begin{tabular}[c]{@{}c@{}}Difficulty vs.\\Brittleness\end{tabular}
 & \cmk
 & \cmk
 & \begin{tabular}[c]{@{}c@{}}Meta-evals wrapper\\ over \texttt{lm-eval-harness}\\ + \includegraphics[height=1em]{files/HF.png} Datasets\end{tabular} \\
\bottomrule
\end{tabular}
}
\caption{
Comparison of \bb{} with recent work on robustness evaluation.
While prior work addresses complementary aspects of prompt robustness, \bb{} brings these dimensions together in a unified framework, combining broad perturbation coverage with variance decomposition, reasoning and ranking-stability analyses, and an open-source meta-evaluation pipeline applicable to both open-weight and frontier API models.}
\label{tab:related_work}
\end{table*}

In this work, we introduce \bb{}, a theoretical and empirical framework for measuring performance differences among models under diverse prompt perturbations.
Our method decomposes observed performance variance into components attributable to task difficulty and prompt sensitivity.
Applying this variance decomposition framework to both frontier and state-of-the-art open-weight models, we find that introducing semantics-preserving input perturbations can decrease average performance by up to $\sim$21\%.
Even among frontier models, Claude Opus 4.5 loses roughly 2\% on MMLU from character-level typos alone, and the variability is large enough to reorder leaderboard rankings.
Our results highlight the substantial impact of prompt sensitivity and underscore the need to more accurately capture this effect within current evaluation frameworks.

\bb{}'s main contributions are the following:

\begin{itemize}

\item We introduce a theoretical variance decomposition framework for quantifying model brittleness, separating performance variability attributable to intrinsic task difficulty from variability induced by prompt formulation.

\item We develop a unified taxonomy of semantics-preserving prompt perturbations that consolidates prior work and introduces several additional perturbation types that affect model performance.

\item We instantiate this framework in \bb{}, a benchmark-agnostic evaluation pipeline that systematically applies perturbations to publicly available benchmarks and evaluates robustness across state-of-the-art open-weight and commercial models.

\item We show that prompt-induced variability persists across tasks, model scales, and reasoning settings, and that perturbations can substantially affect both absolute performance and relative model rankings.
\end{itemize}


\section{Semantics-Preserving Perturbations}
\label{sec:perturbations}

\begin{figure*}
    \centering
    \includegraphics[width=1\linewidth]{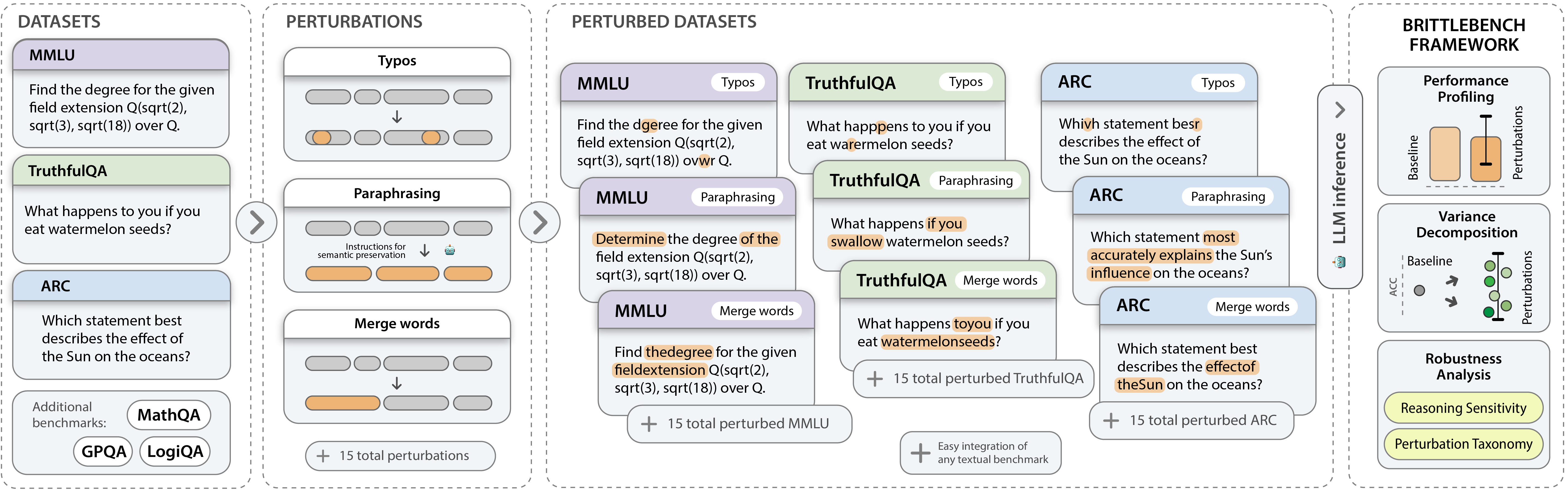} 
    \caption{The \bb{} meta-evaluation framework. We select widely used benchmarks and apply semantics-preserving perturbations as described in Section \ref{sec:perturbations}. We evaluate both the original and perturbed benchmarks using frontier and open-weight state-of-the-art models. 
    \textbf{Evaluating models on \bb's perturbations provides a more comprehensive assessment by accounting for biases introduced by variations in input formats.}}
    \label{fig:framework}
\end{figure*}

To accurately assess model sensitivity to prompt perturbations, and to disentangle it from other sources of performance variance such as task difficulty, we propose a meta-evaluation framework that systematically applies a set of semantics-preserving perturbations to input prompts and measures the resulting performance differentials (Figure \ref{fig:framework}). Our framework adopts a 2-part perspective: 
From the model standpoint, it enables a more faithful estimation of underlying model ability and helpfulness by isolating performance variation attributable to superficial input noise. 
From the benchmark standpoint, it facilitates an assessment of construct validity by examining whether observed difficulty arises from inherent task complexity or from incidental, structural and/or formatting artifacts in the prompt design.

This study contributes to the existing prompt-sensitivity literature in three ways. First, we introduce a unified taxonomy that categorizes all perturbations, thereby enabling a stratified analysis of the performance variance introduced by each perturbation type across different models. Building on previous research in prompt structure analysis \citep{habba2025dove, wu2025rewordbench}, we group perturbations into four categories based on the component of the prompt they affect: \textit{word manipulation, prompt augmentation, prompt padding,} and \textit{paraphrasing perturbations}. \textbf{Word manipulation} encompasses character- or word-level changes such as typos, word splits or merges, and the insertion of additional spaces. \textbf{Context augmentation} involves the addition of information such as personas, emotional phrases, or explanatory paraphrasing. \textbf{Prompt padding} refers to the inclusion of special characters, such as spaces, quotation marks, or newlines, at the beginning and end of prompts. \textbf{Paraphrasing} includes three semantics-preserving rewordings: lexical, syntactic, and rule-free.
A comprehensive explanation of each perturbation type, and a visual summary are in Appendix~\ref{sec:app_perturbations}.

Second, we expand the space of perturbation variants explored in existing literature, for example, by examining prompt padding with a broader range of special characters beyond commonly used quotation marks.
Our methodology leverages a diverse set of perturbations established in the literature, including query prompt augmentation \citep{li2024evaluating, sinha2023language, li2023large}, format changes \citep{habba2025dove, agrawal2025enhancing, zhu2306promptrobust}, context paraphrasing \citep{cao2024worst, deng2023rephrase}, and input padding with special characters \citep{wu2025rewordbench}.
Third, we conduct a fine-grained sensitivity analysis on specific perturbations, with a particular focus on typos, paraphrasing, and prompt padding (see \cref{fig:sensitivity}), which shows that introducing more perturbations of a particular type worsens performance.

We validate semantic preservation using a task-centered criterion: a perturbation is considered \textbf{semantics-preserving} when it does not change the expected gold answer or remove information needed to answer the question. 
We evaluate all perturbation types with an LLM judge, sampling 100 instances per benchmark-perturbation pair. 
For the perturbations most likely to affect meaning or answerability (typos, stopword deletion, word merging, and word splitting), three expert annotators additionally label 35 samples per benchmark--perturbation pair, enabling us to measure both inter-annotator agreement and human--LLM judge agreement. 
Overall, 99.8\% of judged cases preserve all crucial information. These results support the assumption that \bb's perturbations primarily induce surface form variation rather than changes in task semantics. Full details are provided in Appendix~\ref{sec:semantic}.


\section{\bb}
\label{sec:methodology}

\bb{} is a framework that applies the taxonomy of semantics-preserving perturbations described above on existing benchmarks, measuring model robustness by decomposing the variance of prompt sensitivity and the inherent task complexity. 
To do so, we use a \textbf{variance decomposition methodology} with a set of proposed metrics suited for quantifying model and benchmark brittleness.

\textbf{Variance Decomposition:}
Standard benchmark evaluation reports a single accuracy number from a fixed prompt formulation, conflating two sources of variability: \emph{item difficulty}, where some items are consistently harder for a model, and \emph{surface-form sensitivity}, where the same item yields different outcomes under semantically equivalent prompt variants. 
We separate these effects using a variance decomposition framework, similarly to \citet{wang2025measuring}.

Let $Y_{ij}\in\{0,1\}$ denote whether the model's generated answer to item $i$ under perturbation condition $j$ is judged correct. For each model--benchmark pair, this gives an item-by-perturbation matrix of observed correctness outcomes. We decompose its variance as
\[
\mathrm{Var}(Y)
=
\underbrace{\mathrm{Var}_D\!\left(\mathbb{E}_P[Y\mid D]\right)}_{V_{\mathrm{item}}}
+
\underbrace{\mathbb{E}_D\!\left[\mathrm{Var}_P(Y\mid D)\right]}_{V_{\mathrm{prompt}}}.
\]
The first term captures stable, model-conditional item difficulty, corresponding to items that are consistently answered correctly or incorrectly across perturbations, while the second captures prompt brittleness, corresponding to cases where the same semantic item changes correctness under surface-form perturbations. 
Therefore, $V_{\mathrm{prompt}}$ measures correctness-level instability under semantics-preserving prompt variation.
Since we use a single inference run, inference-level stochasticity cannot be separated from prompt-induced variation. 
We therefore interpret $V_{\mathrm{prompt}}$ as observed within-item correctness variability under perturbations, and as an upper bound on variance attributable purely to perturbations under stochastic generation.
A detailed derivation of the variance decomposition is provided in Appendix~\ref{app:variance}.

\paragraph{Brittleness scores:} 
In this notation, $D_{mb}$ and $P_{mb}$ are the empirical estimates of $V_{\mathrm{item}}$ and $V_{\mathrm{prompt}}$, respectively, computed separately for model $m$ on benchmark $b$.
For each model--benchmark pair $(m,b)$, the decomposition partitions the total observed binary correctness variance $V_{mb}=\mathrm{Var}(Y)$ into an item component $D_{mb}$ and a prompt component $P_{mb}$, with $V_{mb}=D_{mb}+P_{mb}$.
We define brittleness as the fraction of total observed correctness variance attributable to prompt perturbations. 
We aggregate variance components across benchmarks for each model and across models for each benchmark before normalization, yielding the model-level ($\Pi_m$) and benchmark-level ($\Pi_b$) brittleness scores:
\[
\Pi_m \;=\; \frac{\sum_b P_{mb}}{\sum_b V_{mb}}, \qquad
\Pi_b \;=\; \frac{\sum_m P_{mb}}{\sum_m V_{mb}}.
\]

Higher scores indicate that a larger share of observed correctness variability comes from within-item sensitivity to prompt variants, rather than from stable item-level difficulty. 
At the model level, this means the same items more often flip correctness under perturbations. 
At the benchmark level, this means that observed performance variability on that benchmark is more strongly associated with prompt sensitivity than with consistently easy or hard items.

\section{Experimental Setup}
\label{sec:experiments}

In principle, \bb{} can be applied to any text-based benchmark.  In our experiments, we apply \bb{} perturbations (Section \ref{sec:perturbations}) to a variety of commonly used benchmarks to demonstrate their utility. 
More specifically, we test the multiple-choice-questions (MCQ) benchmarks of MMLU \citep{hendryckstest2021}, TruthfulQA \citep{lin2021truthfulqa}, ARC \citep{allenai:arc}, MathQA \citep{amini-etal-2019-mathqa}, LogiQA \citep{liu2020logiqa}, and GPQA \citep{rein2024gpqa}, and the open-generation tasks HumanEval \citep{chen2021codex} and GSM8K \citep{cobbe2021gsm8k}. 
\bb{} is built on top of the \texttt{lm-evaluation-harness} \citep{eval-harness}, a widely used evaluation engine that underpins large-scale benchmarking efforts such as the HuggingFace Open LLM leaderboard \citep{open-llm-leaderboard-v2}. 
\bb{} extends the harness by systematically applying input perturbations. 
We benchmark the commercial models of GPT-5\footnote{GPT-5 version: \texttt{gpt-5-2025-08-07}} and Claude Opus  4.5\footnote{\href{https://www.anthropic.com/claude-opus-4-5-system-card}{Claude Opus 4.5 system card}}, and state-of-the-art open-weight model families of Llama3.1, Llama3.3 \citep{dubey2024llama} and Qwen3 \citep{yang2025qwen3}. 
For open-weight and commercial models on multiple-choice benchmarks, we prompt the model to output the answer letter (detailed prompt in the Appendix \ref{app:prompt}). 
Open-weight model inference was run on an internal cluster of NVIDIA H200 GPUs, with individual runs taking approximately 5--15 minutes on a single GPU, for an estimated total of 300 GPU-hours.
We run Claude Opus 4.5 with zero temperature and GPT-5 with minimal internal reasoning enabled.
In total, we performed over 1,800 inference runs to support the experiments and analyses presented in this work. 
Details on the perturbation implementations and default settings are provided in Appendix~\ref{app:default-settings}.
Reproducibility details are provided in Section~\ref{sec:reprod}. 
Finally, for paraphrasing-based perturbations, we use GPT-4o \citep{achiam2023gpt} to generate paraphrases of each input question and evaluate the resulting outputs using both human annotation and an LLM-based judge with GPT-4o.
Additional details on paraphrasing perturbations are in the Appendix~\ref{sec:semantic}.

\section{Results \& Analysis}
\label{sec:analysis}

To understand how prompt perturbations affect large language models, we first examine the direct impact of perturbations on model performance and rankings, then decompose performance variability to attribute variance to model brittleness versus task difficulty, and finally study whether test-time inference strategies can mitigate perturbation-induced sensitivity.

\subsection{Impact of perturbations on model performance}

\paragraph{All models and benchmarks exhibit sensitivity to perturbations; surface-form changes have the largest impact.} Across all models and evaluation settings, the impact of input perturbations depends strongly on which part of the prompt they target (Table~\ref{tab:main_results}). 
Word-level manipulations, such as typos or spacing changes, consistently degrade performance, with accuracy drops of several points, even for the largest models we test. 
This indicates that surface-form noise remains a persistent challenge that is not fully mitigated by scale alone.
Prompt padding also leads to systematic degradation, particularly in few-shot settings, with up to 27.34\% performance degradation, as shown in Table~\ref{tab:res_padding} in Appendix~\ref{sec:results}. 
In contrast, LLM-generated paraphrasing perturbations are comparatively benign and occasionally improve performance in open-weight models. This fact can be interpreted in two ways: perhaps models are genuinely more robust to less constrained types of perturbations. 
Alternatively, however, it could simply be that using an LLM to paraphrase might make queries easier for the model, by standardizing language and decreasing diversity \citep{kleinberg-raghavan-2021-algorithmic, padmakumar-he-2024-writing} or by leveraging the model's preference for model-generated text.\footnote{We evaluate LLMs as answer-generators, not LLM-as-judges. However, the fact that performance on LLM-generated paraphrases remains approximately the same might nevertheless be the result of a generalized form of the so-called ``self-bias'' (whereby an LLM-as-judge prefers its own generations; \citealt{panickssery-2024-llm, xu-etal-2024-pride, spiliopoulou-etal-2025-play}).} 
We further find that these trends extend beyond the model families used in our main evaluation, with Phi-4 \citep{abdin2024phi}, Gemma 3 \citep{team2025gemma}, and Mistral \citep{liu2026ministral} exhibiting qualitatively similar sensitivity patterns across a targeted subset of benchmarks and perturbations. Additional results are provided in Appendix~\ref{sec:extra-models}.

\textbf{Commercial frontier models are similarly susceptible to input perturbations.}
Claude Opus 4.5 shows performance drops comparable to open-weight models that are observed under word-level and context-augmentation perturbations, while remaining relatively robust to prompt-padding perturbations. 
In contrast, GPT-5 (minimal thinking mode) exhibits the largest absolute performance declines, despite achieving high overall baseline accuracy. 
Notably, due to computational constraints, many evaluation pipelines deploy GPT-5 in minimal thinking mode; our results therefore reflect a lower bound on its performance under perturbations. 
Overall, despite their superior average performance compared to open-weight models, commercial frontier models remain sensitive to input perturbations, incurring absolute performance drops comparable in magnitude to those observed for open-weight models.

\begin{table*}
  \resizebox{1\textwidth}{!}{%
\begin{tabular}{l|c>{\columncolor[gray]{0.95}}c|c>{\columncolor[gray]{0.95}}c|c>{\columncolor[gray]{0.95}}c|c>{\columncolor[gray]{0.95}}c|c>{\columncolor[gray]{0.95}}cc}
\toprule
\textbf{Model} & \multicolumn{2}{c}{\textbf{\begin{tabular}[c]{@{}c@{}}Word-level\\ Perturbations\end{tabular}}} & \multicolumn{2}{c}{\textbf{\begin{tabular}[c]{@{}c@{}}Prompt\\ Padding\end{tabular}}} & \multicolumn{2}{c}{\textbf{\begin{tabular}[c]{@{}c@{}}Prompt\\ Augmentation\end{tabular}}} & \multicolumn{2}{c}{\textbf{Paraphrasing}} &  \multicolumn{2}{c}{\textbf{Average}} \\
               & \multicolumn{1}{c}{$\Delta Drop$ ($\downarrow$)} & \multicolumn{1}{c}{$Range$} 
               & \multicolumn{1}{c}{$\Delta Drop$ ($\downarrow$)} & \multicolumn{1}{l}{$Range$} 
               &  \multicolumn{1}{c}{$\Delta Drop$ ($\downarrow$)} & \multicolumn{1}{c}{ $Range$}
               & \multicolumn{1}{c}{$\Delta Drop$ ($\downarrow$)} & \multicolumn{1}{c}{$Range$}
               & \multicolumn{1}{c}{$\Delta Drop$ ($\downarrow$)} & \multicolumn{1}{c}{$Range$}              

\\ 
\midrule
\multicolumn{11}{c}{\textit{\textit{Commercial Frontier Models}}} \\

Claude Opus 4.5 &  6.49 & 6.49 & 1.87 & 3.32 & 1.56 & 1.56 & 0.38 & 0.44 & 2.58 & 2.95 \\
GPT-5 & 9.32 & 9.32 & 0.77 & 1.60 & 4.75 & 4.75 & \textbf{5.66} & 5.66 & 5.13 & 5.33 \\

\midrule
\multicolumn{11}{c}{\textit{\textit{Open-Weight Models}}} \\

Qwen-32b & 7.54 & 7.54 & 5.41 & 5.41 & 0.83 & 0.83 & 0.49 & 0.49 & 3.57 & 3.57 \\
Qwen-8b & 8.50 & 8.52 & 16.47 & 16.51 & 0.95 & 0.95 & 0.60 & 0.60 & 6.63 & 6.65 \\
Qwen-4b &  9.27 & 9.27 & 7.17 & 7.60 & 0.79 & 0.79 & 1.18 & 1.18 &  4.60 & 4.71 \\
Llama3.3-70b &  7.49 & 7.55 & 6.65 & 6.98 & 1.88 & 1.89 & 0.88 & 0.88 & 4.23 & 4.33 \\
Llama3.1-70b &  7.08 & 7.08 & 10.51 & 10.51 & 5.31 & 7.45 & -0.01 & 2.37 & 5.72 & 6.85 \\
Llama3.1-8b &  \textbf{9.42} & 9.42 & \textbf{21.61} & 21.61 & \textbf{14.78} & 16.59 & 0.00 & 3.23 & \textbf{11.45} & 12.71 \\
 
\\ \bottomrule
\end{tabular}
}
\caption{Mean $\Delta Drop$ and $Range$ (\%) per model and perturbation category across the six MCQ benchmarks used for \bb{}: \textit{MMLU, ARC, TruthfulQA, MathQA, GPQA, and LogiQA}. Perturbations are grouped by the type of change they make to the input prompt. For each perturbation group, $\Delta Drop$ denotes the largest drop from the clean baseline to the lowest-performing perturbation in the group, and $Range$ denotes the difference between the best and worst perturbation performance within the group. Each pair of values reports these quantities averaged across the six benchmarks in the 0-shot setting. Disaggregated results for each perturbation group are provided in Appendix~\ref{sec:results}. Confidence intervals for the reported aggregate values are provided in Appendix~\ref{sec:uncertainty}.}
  \label{tab:main_results}
\end{table*}

\begin{figure*}
    \centering
    \centerline{\includegraphics[width=0.95\columnwidth]{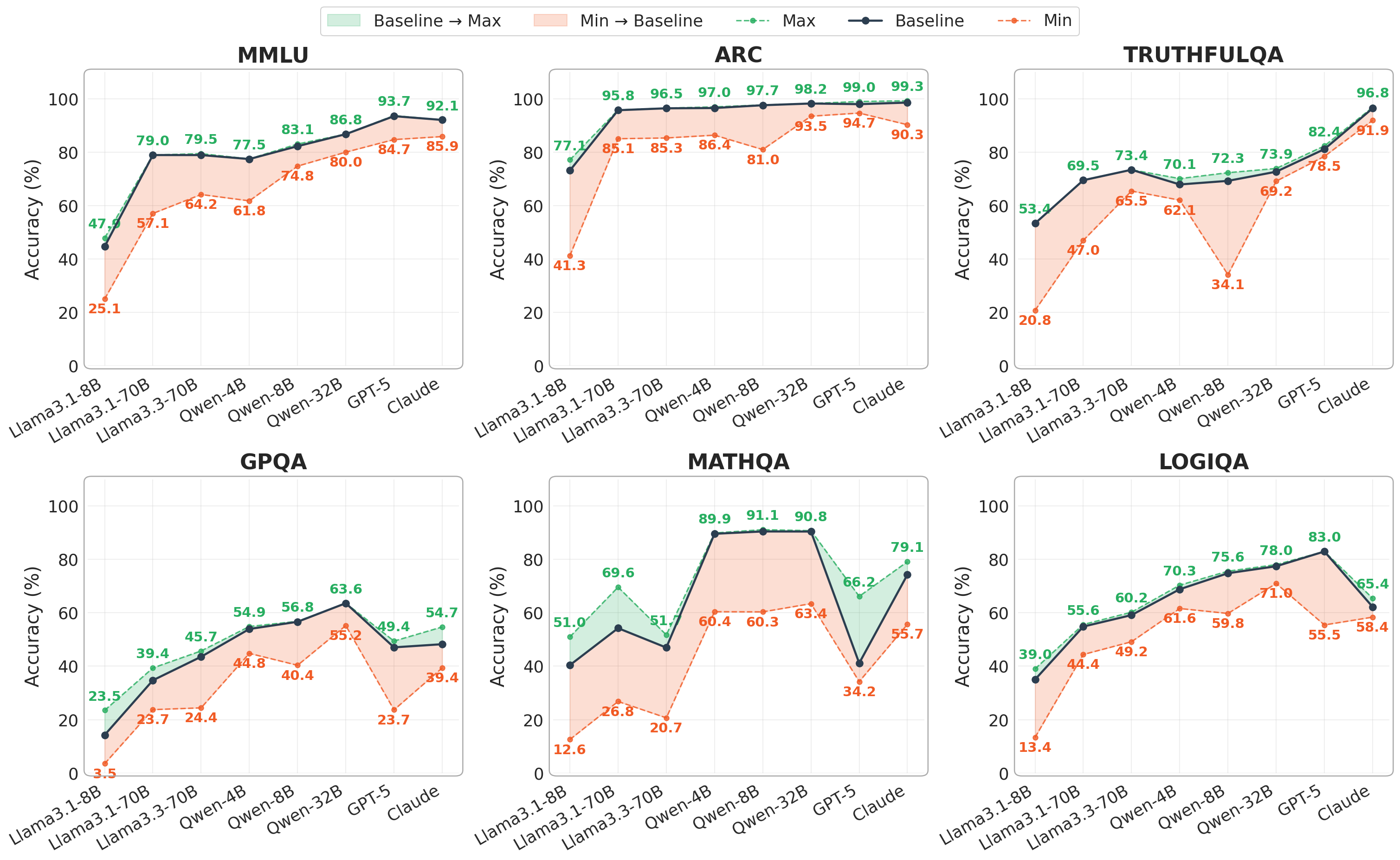}}
    \caption{\textbf{Input perturbations systematically increase performance variance across models.} Line plots show Accuracy (\%) of each model on six benchmarks under zero-shot evaluation. The solid line shows baseline performance; dashed lines indicate the best- and worst-case accuracy across all perturbation types, with shaded bands highlighting the gap between   
  the baseline and each extreme.}
    \label{fig:results}
\end{figure*}

\paragraph{Model rankings are sensitive to prompt perturbations.}
Beyond absolute performance degradation (Figure~\ref{fig:results} \& Table~\ref{tab:main_results}), benchmarks are commonly used for leaderboarding, and perturbations can alter model rankings by affecting models unevenly.
We quantify this effect by computing model rankings for the six open-weight models on each task before and after applying a perturbation. 
We measure their agreement using Spearman rank correlation, averaged across tasks and reported separately for zero-shot and few-shot settings (Table~\ref{tab:rank_correlation}).
Overall, introducing even a single perturbation changes the relative ranking of the six open-weight models in 63\% of cases, indicating that model comparisons are often sensitive to prompt formulation. The magnitude of this effect depends strongly on the type of perturbation. 
In both zero-shot and few-shot settings, \textit{prompt padding} with quotation marks and newlines yields the lowest rank correlations (zero-shot: $0.63-0.67$; few-shot: $0.73$), frequently altering model rankings compared to the baseline. 
In contrast, \textit{adding spaces around punctuation} or \textit{splitting words} produces much higher rank correlations (zero-shot: $0.87-0.91$; few-shot: $0.95-0.96$), indicating minimal impact on relative ordering. 
Rankings are generally more stable in the few-shot setting, however, sensitivity remains non-trivial, as even the least impactful perturbations still change rankings in roughly half of the evaluated tasks.

\textbf{Even simple surface-form perturbations induce nonlinear degradation as their prevalence increases.} We examine how model performance varies with the prevalence of common surface-form perturbations.
Specifically, we vary the number of instances of typos and prompt-padding perturbations (spaces, quotation marks, and newlines) injected into each prompt and measure the resulting performance on two large-scale open-weight models, Llama3.3-70B and Qwen3-32B, across all benchmarks. 
As the number of perturbations increases, both models exhibit non-uniform degradation patterns across perturbation types (Figure \ref{fig:sensitivity}a). 
Typos lead to a largely monotonic decline in performance, while, in contrast, prompt-padding perturbations show more irregular behavior. Spaces and new lines are comparatively benign and often result in stable or mildly fluctuating performance, while quotation marks induce sharper drops at higher intensities. 
Overall, our results demonstrate that robustness is not only perturbation-type dependent but also intensity-dependent, highlighting the importance of evaluating models beyond single-instance perturbations and toward more realistic, compounded noise settings.

\subsection{Variance and Brittleness Attribution}
\label{sec:variance_analysis}

\paragraph{Perturbed benchmarks reveal richer information about model performance than unperturbed benchmarks.}  Evaluating models on perturbed benchmarks reveals substantially richer performance variability than evaluation on the original benchmarks alone (Figure~\ref{fig:results}).
While baseline benchmarks produce relatively compact performance distributions, the perturbed counterparts expose a wider spread in accuracy, uncovering sensitivity that is otherwise masked by single-format evaluations. 
Perturbations expose additional variability across models and scales, indicating that robustness to input variations is not uniform even when average performance appears stable.
Consequently, relying solely on original benchmarks can overestimate model reliability, whereas incorporating perturbed evaluations enables a more comprehensive characterization of model behavior under realistic input conditions. 
These results underscore the importance of robustness-aware evaluation frameworks that account for variation in input formats rather than treating benchmark performance as a single-point estimate.

\begin{figure*}
    \centering
    \centerline{\includegraphics[width=\columnwidth]{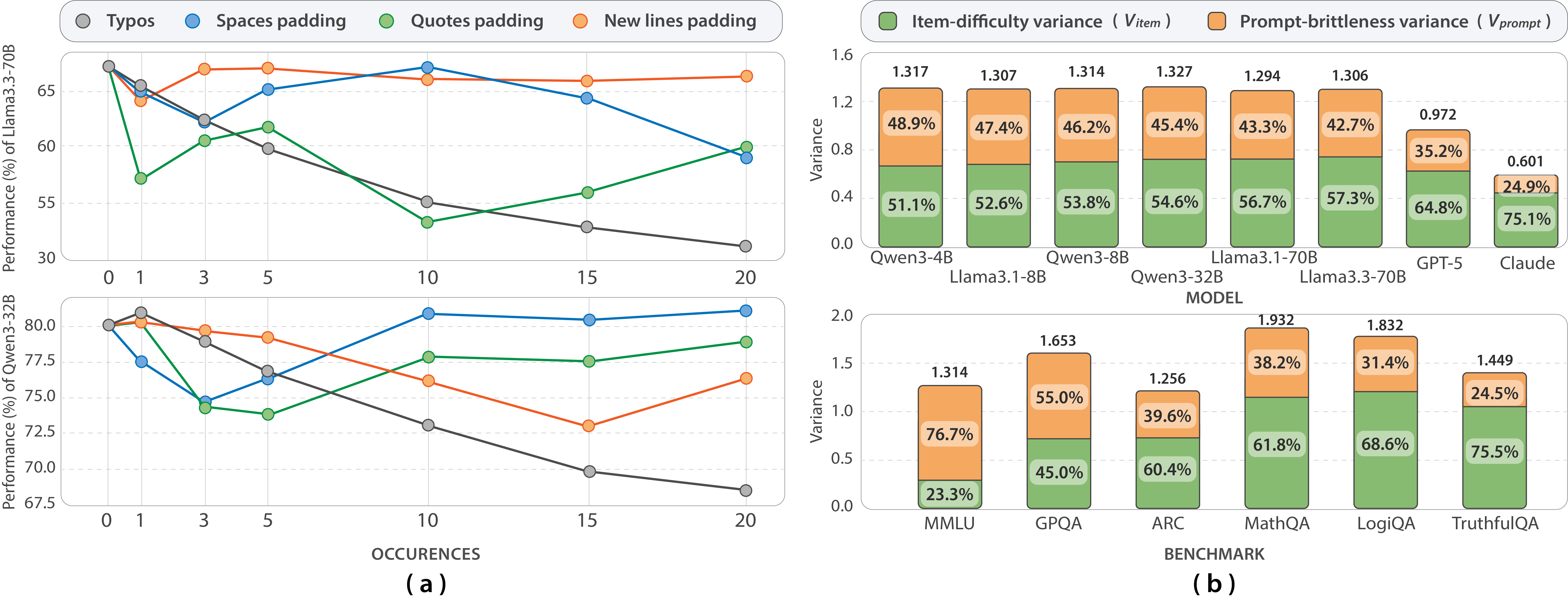}}
    \caption{\textbf{(a): Model sensitivity to perturbation intensity.}  Llama3.3-70B \textit{(top)} and Qwen3-32B \textit{(bottom)} accuracy (\%) aggregated across benchmarks, under increasing numbers of surface-form perturbations, including typos and prompt-padding variations. Each line shows accuracy as a function of the number of perturbation occurrences, illustrating how perturbation intensity affects robustness. \textbf{(b): Variance decomposition} for models \textit{(top)} and benchmarks \textit{(bottom)}. From the model's perspective, the \textit{brittleness score} ($\Pi_m$) quantifies whether a model’s performance variability across tasks is dominated by prompt sensitivity rather than intrinsic item difficulty. From the benchmark perspective, the \textit{brittleness score} ($\Pi_b$) indicates whether a benchmark primarily discriminates models based on task difficulty or on sensitivity to input perturbations.}
    \label{fig:sensitivity}
\end{figure*}

\textbf{Perturbation-induced variability accounts for a substantial fraction of total performance variance.} The model-level brittleness scores (Figure~\ref{fig:sensitivity}b -- \textit{top}) reveal that, across all evaluated models, perturbation-induced variance constitutes a substantial fraction of total performance variability, accounting for roughly half of the observed variance in most open-weight models. Notably, this pattern is consistent across model scales: both larger models (e.g., Llama3.3-70B, Llama3.1-70B) and smaller models (e.g., Qwen3-4B, Llama3.1-8B) exhibit perturbation-driven variance comparable to, or exceeding, variance attributable to item difficulty. 
Commercial models display lower absolute variance overall, yet prompt sensitivity still accounts for more than $\sim25\%$ of their total variance. 
Consequently, differences in model performance across tasks are not solely governed by intrinsic task difficulty but are strongly influenced by robustness to input variations. These findings underscore that robustness is a distinct axis of model behavior that cannot be inferred from mean performance alone.

\textbf{Benchmarks differ substantially in whether model performance variability is driven by task difficulty or by sensitivity to input perturbations.} Benchmark-level brittleness scores (Figure \ref{fig:sensitivity}b - \textit{bottom}) demonstrate pronounced heterogeneity in how benchmarks distribute performance variance between item difficulty and perturbation sensitivity. 
MMLU and GPQA are dominated by perturbation-induced variance, with over half of their total variance attributable to sensitivity to input perturbations rather than underlying task difficulty. 
TruthfulQA, LogiQA, MathQA, and ARC show variance profiles largely driven by item difficulty, indicating stronger measurement stability under input perturbations.
For the brittleness-dominant benchmarks, this suggests that many models have already reached similar levels of semantic competence on these tasks, causing surface-form robustness to become the main factor distinguishing them. 
Importantly, this does not indicate a flaw in the benchmark itself; instead, it reflects a saturation effect under modern models.
As a consequence, benchmark scores can increasingly capture how robust a model is to input variations unless perturbation-aware evaluations are explicitly incorporated.

\begin{figure*}
    \centering
    \includegraphics[width=1\linewidth]{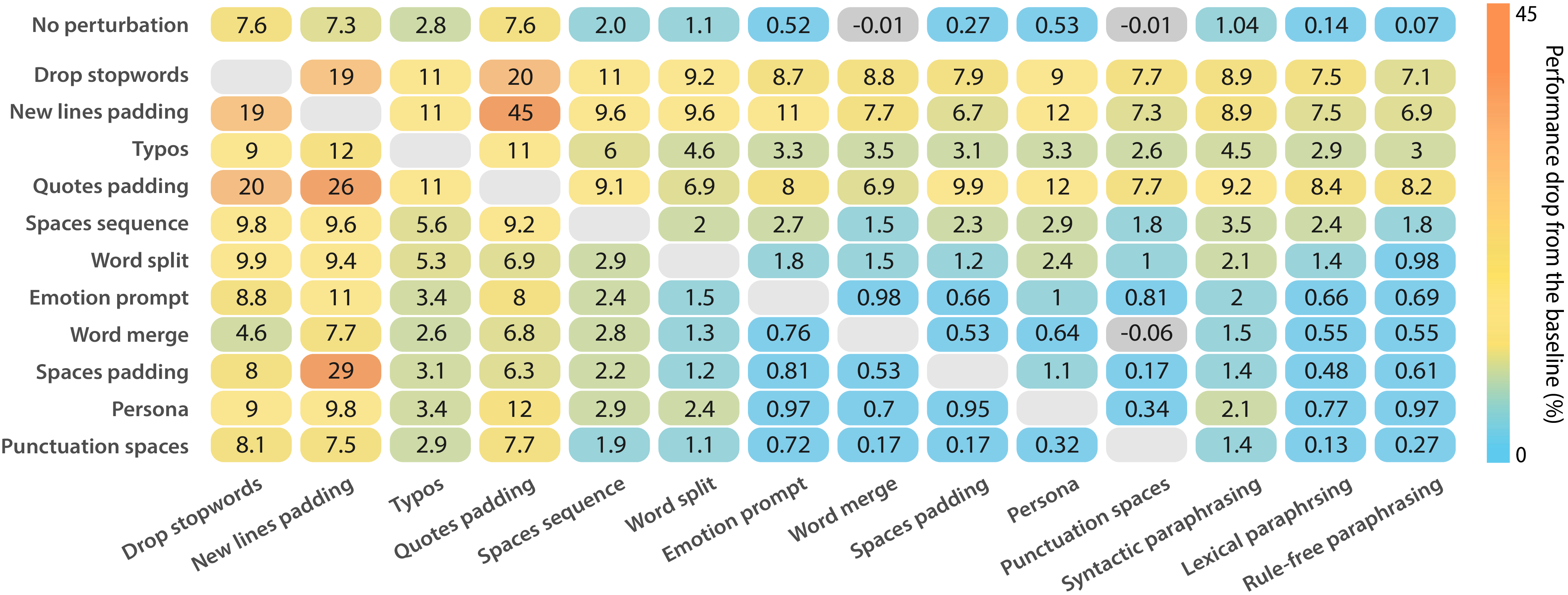} 
    \caption{Heatmap of Accuracy Drop (in \%; $\downarrow$) from the baseline on MMLU for Qwen3-8B (0-shot) under single and paired input perturbations of \bb{}. The first row reports the performance drop for individual perturbations, while each row shows the additional effect of applying a second perturbation on top of the first. Paraphrasing perturbations are always applied once. Higher values indicate larger degradation, revealing both compounding and counteracting effects between perturbations. Perturbations are sorted based on their average performance drop on all the benchmarks and models from highest (left/top) to lowest (right/bottom).}
    \label{fig:stacked}
\end{figure*}

\subsection{Inference-time robustness analysis}
\label{sec:icl}

\textbf{Compositional input perturbations substantially exacerbate model brittleness.} 
We apply pairs of perturbations sequentially to Qwen3-8B on MMLU as a case study, and measure the resulting change in accuracy (performed to ensure no overwriting of the first perturbation by the second). Combining perturbations generally amplifies performance degradation (Figure~\ref{fig:stacked}), with some combinations leading to drops of up to $\sim45\%$, far exceeding the effect of individual perturbations. Such drops also exceed the absolute differences between models' baseline performances ($\approx$~2--10\% in our sample). Notably, the interaction between perturbations is order-dependent, as evidenced by the asymmetry around the diagonal of the heatmap, indicating that the same, cumulative perturbations applied in different orders can yield substantially different outcomes. Finally, we observe little evidence of systematic mitigation effects: combining perturbations rarely offsets performance loss, and in most cases further degrades model accuracy rather than stabilizing it.

\textbf{Chain-of-thought (CoT) boosts accuracy but only weakly mitigates brittleness to prompt perturbations.}
We evaluate CoT reasoning with Claude Opus 4.5 across six benchmarks and the most impactful word-level perturbations: \textit{typos}, \textit{stopword deletion}, and \textit{insertion of extra spaces between words}. As shown in Appendix~\ref{sec:app_reasoning}, CoT substantially outperforms non-CoT prompting, producing correct answers in 2.4× more cases than those in which it introduces new errors (1.7\% error rate with CoT versus 4\% without). 
However, these gains are uneven across benchmarks, with limited improvements on tasks such as MathQA and TruthfulQA. 
Moreover, higher accuracy does not translate into commensurate robustness: under perturbations, CoT reduces the accuracy drop by only 0.41 percentage points (from 2.79\% to 2.38\%), corresponding to a 17\% relative reduction.
Beyond reducing accuracy, perturbations also consistently degrade the quality of reasoning traces, as assessed by an LLM-based judge (GPT-5; Appendix \ref{sec:app_reasoning}), and higher reasoning quality is associated with higher accuracy. Taken together, these results suggest that reasoning can mitigate brittleness only when the reasoning process itself remains intact.

\section{Related Work}
\label{sec:related_work} 

\textbf{Structural and Lexical Perturbations:}
Prior work suggests LLMs are often not robust to semantically-preserving structural perturbations. \citet{gu-etal-2023-robustness}, \citet{pezeshkpour-hruschka-2024-large}, and \citet{haller2025llm} study robustness to a diverse set of lexical prompt perturbations.
\citet{sclar2024quantifying} shows that even minor perturbations to formatting, such as changing the casing, spaces, or separators in the prompt, can have significant impacts on model performance. 
\citealt{ross-etal-2022-tailor} offers bespoke tooling to generate structural paraphrases with semantic controls. 
Multiple-choice option ordering has also been shown to affect model performance \citep{gupta-etal-2024-changinganswerorderdecrease, alzahrani2024benchmarks}.
PromptRobust \citep{zhu2306promptrobust} is a benchmark designed to evaluate model robustness, combining aspects of both structural and stylistic perturbations from prior work.
Other evaluation suites aggregate multiple perturbation types to assess robustness across different dimensions \citep{wang2021textflint, habba2025dove, mizrahi-etal-2024-state}.

\textbf{Stylistic and Paraphrased Perturbations:}
\citet{sun2024evaluating} systematically evaluate prompt sensitivity to human-written, paraphrased instructions over MMLU \citep{hendryckstest2021} and BBL \citep{srivastava2023beyond}, and find consistent performance degradation on the paraphrased instructions. 
Similarly, \citet{zhuo-etal-2024-prosa} propose stylistic perturbations to compute an instance-level sensitivity score, using a framework that introduces changes to output requirements, role-playing instructions, emotional framing, and simplifications of the input.
\citet{cao2024worst} propose a benchmark consisting of semantically equivalent queries (paraphrased by an LLM) and argue that worst-case performance across prompt variants is a lower bound on model capability that is more informative than simple average performance.
Other approaches propose scalar indices to quantify prompt sensitivity \citep{chatterjee-etal-2024-posix}, although these metrics are typically tied to specific perturbation families.
Paraphrase-based perturbations mimic expected variability in real-world interactions with LLMs, and performance sensitivity raises concerns about the validity of model performance over popular benchmarks.

As summarized in Table \ref{tab:related_work}, \bb{} combines a broad perturbation coverage with a variance decomposition methodology that isolates brittleness from item difficulty, alongside reasoning- and ranking-stability analyses across frontier and open-weight models.
\section{Conclusion}

In this work, we introduce \bb{}, an evaluation framework for quantifying model brittleness that disentangles task-induced difficulty from performance variability due to prompt sensitivity. We operationalize brittleness through a unified taxonomy of semantics-preserving perturbations and instantiate this approach in an end-to-end, benchmark-agnostic meta-evaluation pipeline. Using \bb{}, we evaluate both state-of-the-art open-weight and commercial models, enabling systematic comparisons across tasks, perturbation types, reasoning settings, and model scales. Our results show that performance on widely used benchmarks can degrade by up to $\sim$21\% under meaning-preserving input perturbations, and that such perturbations can account for up to half of total performance variance. These findings underscore the limitations of current evaluation practices and motivate the adoption of robustness-aware evaluation frameworks to more faithfully assess model capability.

\section*{Acknowledgments}
We shared early versions of this work with several of our colleagues and would like to thank them for their input and suggestions. Thank you: Levent Sagun, Julian Michael, Shivam Singhal, Mark Tygert, Miles Turpin, Marjan Ghazvininejad, Hunter Lang, Richard Pang, Maryam Fazel-Zarandi. Thanks as well to Leshem Chosen for comments on an early draft.

\section*{Reproducibility statement}
\label{sec:reprod}

To support reproducibility, we implement \bb{} as a lightweight wrapper around the open-source \texttt{lm-evaluation-harness} evaluation framework. The wrapper integrates seamlessly with existing harness tasks: for any implemented text-based benchmark, it applies the specified \bb{} perturbations and then runs inference according to the corresponding task configuration and evaluation protocol. Our code, together with detailed instructions for installing the wrapper, generating perturbations, and reproducing the evaluation runs, is available online. 
\footnote{\url{https://github.com/facebookresearch/Brittlebench}}.


\clearpage
\newpage
\bibliographystyle{assets/plainnat}
\bibliography{paper}

\clearpage
\newpage
\beginappendix

\section{Perturbations implementation details}
\label{sec:app_perturbations}

\subsection{Meta-evaluation system design}


To provide a comprehensive model assessment, \bb{} integrates flexible and scalable perturbation capabilities, allowing for the easy addition of custom perturbations, while also using a large pool of existing ones and benchmarks, with built-in support for robustness analysis and similarity metrics for each perturbation. Our approach also provides universal model and evaluation support, making it compatible with any model architecture and enabling both probabilistic and generative evaluation, all optimized for efficient execution, even with vLLM. The efficient and user-friendly orchestration further simplifies the evaluation workflow, maximizing throughput with advanced caching and profiling and allowing for single-call or single-script orchestration.

\begin{figure*}[h] 
    \centering
    \includegraphics[width=1\linewidth]{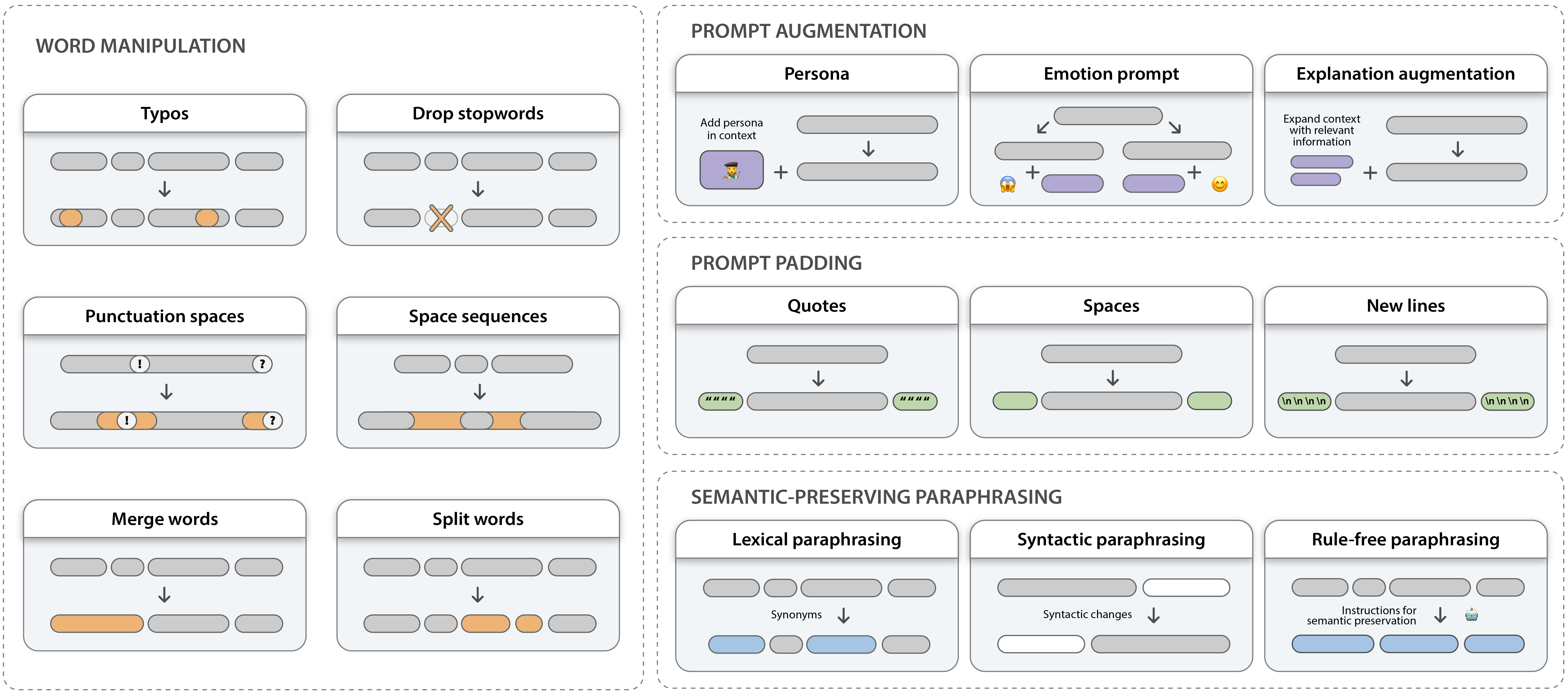} 
    \caption{Visual explanation of perturbations used in \bb{}. Perturbations are grouped based on their type as described in Section \ref{sec:perturbations}.}
    \label{fig:perturbations}
\end{figure*}

\subsection{Word-level perturbations settings}
\label{app:default-settings}
Table \ref{tab:perturbation-implementation} summarizes the implementation details for the word-level perturbations with configurable settings, including the library or function used, its configurable parameters, and the default intensity used in our main experiments. Most word-level perturbations are implemented using AugLy \citep{papakipos-etal-2022-augly}. For the main results in Table \ref{tab:main_results} and Figure \ref{fig:results}, we apply each perturbation once using these default settings.

\begin{table*}[t]
\centering
\footnotesize
\renewcommand{\arraystretch}{1.15}
\setlength{\tabcolsep}{4pt}

\begin{tabularx}{\textwidth}{
    >{\raggedright\arraybackslash}p{0.15\textwidth}
    >{\raggedright\arraybackslash}X
    >{\raggedright\arraybackslash}p{0.35\textwidth}
}
\toprule
\textbf{Perturbation} &
\textbf{Parameters} &
\textbf{Default setting} \\
\midrule

\textbf{Typos perturbation}
&
\textbf{Package / Implementation}: 
\textbf{\textcolor{blue_agg}{\texttt{simulate\_typos()}}} from \textbf{\textcolor{blue_agg}{\texttt{augly.text}}}.

\textbf{Parameters:}
\textbf{\textcolor{green_agg}{\texttt{aug\_word\_p}}}: Per-word probability of applying a typo.
\textbf{\textcolor{green_agg}{\texttt{aug\_char\_p}}}: Per-character probability of applying a typo.
\textbf{\textcolor{green_agg}{\texttt{aug\_word\_min}}}: Minimum number of words to perturb per
\textbf{\textcolor{green_agg}{\texttt{simulate\_typos}}} call.
\textbf{\textcolor{green_agg}{\texttt{aug\_word\_max}}}: Maximum number of words to perturb per
\textbf{\textcolor{green_agg}{\texttt{simulate\_typos}}} call.
&
\makecell[tl]{
\texttt{aug\_word\_p: float = 0.5,}\\
\texttt{aug\_char\_p: float = 0.5,}\\
\texttt{aug\_word\_min: int = 1,}\\
\texttt{aug\_word\_max: int = 1000}
}
\\
\midrule

\textbf{Padding perturbations}
&
\textbf{Package / Implementation}: Original implementation.

\textbf{Parameters:}
\textbf{\textcolor{green_agg}{\texttt{char}}}: Character used for padding. Possible values:
\texttt{<space>}, \texttt{"}, or \texttt{<new line>}.
\textbf{\textcolor{green_agg}{\texttt{char\_count}}}: Number of characters added at the beginning and end of the text.
&
\texttt{char\_count: int = 5}
\\
\midrule

\textbf{Add-space-sequence perturbation}
&
\textbf{Package / Implementation}: Original implementation.

\textbf{Parameters:}
\textbf{\textcolor{green_agg}{\texttt{prob}}}: Probability, from 0.0 to 1.0, of adding a space sequence after a given character.
\textbf{\textcolor{green_agg}{\texttt{min\_spaces}}}: Minimum number of spaces to add, inclusive.
\textbf{\textcolor{green_agg}{\texttt{max\_spaces}}}: Maximum number of spaces to add, inclusive.
&
\makecell[tl]{
\texttt{prob: float = 0.1,}\\
\texttt{min\_spaces: int = 1,}\\
\texttt{max\_spaces: int = 10}
}
\\
\midrule

\textbf{Split/merge words perturbations}
&
\textbf{Package / Implementation}: \textbf{\textcolor{blue_agg}{\texttt{merge\_words()}}} and
\textbf{\textcolor{blue_agg}{\texttt{split\_words()}}} from \textbf{\textcolor{blue_agg}{\texttt{augly.text}}}.

\textbf{Parameters:}
\textbf{\textcolor{green_agg}{\texttt{manipulation\_type}}}: Type of manipulation, either
\texttt{"split"} or \texttt{"merge"}.
\textbf{\textcolor{green_agg}{\texttt{aug\_word\_p}}}: Probability of augmenting each word.
\textbf{\textcolor{green_agg}{\texttt{min\_char}}}: Minimum number of characters in a word to be merged.
\textbf{\textcolor{green_agg}{\texttt{aug\_word\_min}}}: Minimum number of words to augment.
\textbf{\textcolor{green_agg}{\texttt{aug\_word\_max}}}: Maximum number of words to augment.
\textbf{\textcolor{green_agg}{\texttt{n}}}: Number of augmentations performed for each text.
&
\makecell[tl]{
\texttt{aug\_word\_p: float = 0.3,}\\
\texttt{min\_char: int = 2,}\\
\texttt{aug\_word\_min: int = 1,}\\
\texttt{aug\_word\_max: int = 1000,}\\
\texttt{n: int = 1}
}
\\
\bottomrule
\end{tabularx}
\caption{Implementation details and default parameter values for Word-level perturbations with configurable settings.}
\label{tab:perturbation-implementation}
\end{table*}

\subsection{Validating semantic equivalence}
\label{sec:semantic}
\textbf{Assessing semantic preservation of word-level perturbations}:
The \bb{} framework relies on perturbations preserving the meaning of the original, by which we mean that the two are equivalent insofar as the expected gold answer does not change (minor variants to style or formality that do not affect the output are acceptable in our framework). In other words, we define ``semantic-preservation'' behaviorally, on the basis of the task at hand. 

This being said, applying word-level perturbations such as \textit{typos}, \textit{word merges}, and \textit{word splits} can, in some cases, alter the surface form of an input question in ways that may inadvertently break semantic equivalence between the original and a perturbed variant. 
To avoid this problem, the perturbations implemented in the \bb{} meta-evaluation framework are designed with careful, test-driven validation to minimize unintended meaning changes. 
For example, when applying typos to benchmarks containing equations or mathematical expressions, perturbations are explicitly restricted from modifying these components.

The \bb{} behavioral definition of semantic equivalence has some consequences. For example, a variant can be deemed semantically-equivalent to the original according to the definition, but a particular model may nevertheless tokenize the two examples completely differently. As one progresses deeper into model internals, further difference arising from formatting variations can be observed \citep{davidson2025do}. In other words, from the model's ``perspective'', the queries are quite different, but from a human perspective, they are equivalent. Moreover, to the human---in particular, to a user in the real world---such a low level modeling decision is irrelevant, because the same output is expected.\footnote{Like \citealt{chatterjee-etal-2024-posix}, our definition relies to some extent on the ``intent'' of a query, but we break with that work building our notion of semantics-preservation on the gold answer, on the model \emph{behavior}, as opposed to its internal processes or representations.}  

To investigate model notions of semantic equivalence as well, we compute similarity metrics between each original input and its perturbed counterpart, for all perturbations except the paraphrasing ones, as an additional analysis. 
Specifically, we encode each sentence pair using Qwen3-Embedding-8B embeddings \cite{qwen3embedding} and measure cosine similarity between the resulting vectors. 
Across all perturbation types, we observe consistently high similarity scores, with an average cosine similarity of 0.962, supporting the semantic preservation of our perturbations from the model ``perspective'', suggesting that our perturbations are not only behaviorally semantics-preserving by design, but are also largely semantics-preserving according to a model's ``perspective'' as well.

\textbf{Human--LLM Judge validation of semantic preservation}: 
We define semantic preservation behaviorally: a perturbed instance is considered semantically equivalent to the original when the expected gold answer remains unchanged and the perturbation does not remove or alter information necessary to answer the question. This definition is aligned with the goal of Brittlebench, which is to measure sensitivity to surface-form variation rather than changes in task content. Because some perturbations are more likely than others to inadvertently affect meaning, we conduct a targeted validation study on the four highest-risk word-level perturbations: typos, stopword deletion, word merging, and word splitting. We exclude perturbations that primarily affect formatting or prompt structure, such as prompt padding, since these do not modify the lexical content of the question.

For each benchmark and perturbation type, we sample 35 instances for expert human annotation. Three expert annotators evaluate each original--perturbed pair using a two-step protocol. First, annotators make a binary judgment of whether the perturbation changes or removes crucial information such that the question can no longer be reliably answered with the same gold answer. Second, for cases judged as altered, annotators categorize the source of the issue into one of three predefined error types: \emph{Negations \& Logical Keywords}, \emph{Math Symbols \& Formulas}, or \emph{Scientific Constants \& Keywords}. These categories were chosen to capture the main ways in which otherwise superficial perturbations can affect answerability, for example, by breaking a logical relation, modifying a mathematical expression, or corrupting a domain-specific keyword.

To scale the analysis beyond the expert-annotated subset, we use an LLM-based judge, GPT-5 with medium reasoning, with the same binary decision and error-categorization protocol. The LLM judge evaluates 100 samples per benchmark--perturbation pair, for a total of 2,400 judged instances across six benchmarks and four perturbation types. Human agreement is high, with quadratically weighted Cohen's $\kappa = 0.714$, Krippendorff's $\alpha = 0.833$, and raw agreement of $0.892$. Agreement between the human annotations and the LLM judge is also strong, with raw agreement of $0.940$, indicating that the semantic-preservation judgment is well-defined and reliably measurable.

Across all benchmarks and perturbation types, the vast majority of perturbed examples preserve the crucial information needed to answer the question. The LLM judge identifies only five failures out of 2,400 samples, corresponding to an overall preservation rate of approximately $99.8\%$. These failures are rare and systematic: they occur when a perturbation affects semantically critical tokens, such as logical expressions, relational words between quantities, mathematical notation, or scientific constants. For example, splitting or modifying phrases such as ``greater than'' can change the logical structure of a question, and deleting relational words such as ``of'' or ``to'' can make a quantitative relation ambiguous. The failures are confined to two benchmarks and two perturbation types, suggesting that they reflect identifiable edge cases rather than a broad failure of the perturbation process.

\textbf{Assessing paraphrasing quality}: 
Paraphrase quality was assessed by three annotators (AI experts) using an ordinal 5-point scale (1 = very poor paraphrase, 5 = excellent paraphrase). Because the paraphrases produced by GPT-4o were generally of high quality, the observed score distribution was strongly skewed toward the upper end of the scale, with approximately 98\% of ratings in the 4--5 range. Such range restriction is known to reduce the stability and interpretability of chance-corrected agreement statistics, despite high substantive agreement among annotators.
To ensure consistent interpretation of the full rating scale and to anchor the lower end of the ordinal judgments, we included a small set of calibration items consisting of deliberately low-similarity paraphrases (approx. 5\% of the samples). These negative samples were designed to be plausible but semantically deficient paraphrases, and were interleaved with the main evaluation items during annotation. Their purpose was not to evaluate system performance, but to reduce scale ambiguity and to verify annotator consistency across the full quality spectrum.
Inter-annotator agreement was quantified using metrics appropriate for ordinal data and skewed score distributions: quadratically weighted Cohen’s $\kappa$ (averaged across annotator pairs) of \textit{0.781} and Krippendorff’s $\alpha$ of \textit{0.836} with an ordinal distance function. These metrics weight disagreements proportionally to their magnitude (e.g., penalizing 4 vs. 5 disagreements substantially less than 1 vs. 5 disagreements) and are therefore robust to imbalanced label distributions. For completeness, we additionally report the raw exact agreement of \textit{0.902}, which is generally considered high.

In addition to human annotation, each paraphrase was evaluated by an LLM-based judge (GPT-4o) using the same 5-point ordinal rubric. To compare the judge with human assessments, we aggregated the three human ratings per item into a single reference score using the mean human rating, and derived a discrete human consensus label by rounding the mean to the nearest integer. Agreement between the LLM judge and the human reference was substantial, with a quadratically weighted Cohen’s $\kappa$ of \textit{0.678}, indicating strong ordinal consistency. This alignment is further reflected by a high raw exact agreement of \textit{0.833} between the judge and the rounded mean human score, and a low mean absolute difference of \textit{0.244} between the judge score and the mean human rating, suggesting that disagreements were generally small in magnitude. Finally, the judge’s scores exhibited a moderate monotonic association with human judgments, as measured by Spearman’s rank correlation of \textit{0.447}, indicating that the judge largely preserved the relative ordering of paraphrase quality as perceived by humans.

\section{Variance Decomposition}
\label{app:variance}

\textbf{Full decomposition derivation:}
Let $Y = Y(D,P,R)$ denote binary correctness, where $D$ indexes the data item, $P$ the perturbation condition, and $R$ the inference run.
We first apply the law of total variance with respect to the inference run:
$$
\mathrm{Var}(Y)
=
\mathrm{Var}_{D,P}\!\left(\mathbb{E}_{R}[Y \mid D,P]\right)
+
\mathbb{E}_{D,P}\!\left[\mathrm{Var}_{R}(Y \mid D,P)\right].
$$

The first term captures systematic variation across items and perturbations after averaging over inference runs. The second term captures run-to-run variability for a fixed item and perturbation.
We then apply the law of total variance to the first term with respect to the data item:
$$
\mathrm{Var}_{D,P}\!\left(\mathbb{E}_{R}[Y \mid D,P]\right)
=
\mathrm{Var}_{D}\!\left(
\mathbb{E}_{P,R}[Y \mid D]
\right)
+
\mathbb{E}_{D}\!\left[
\mathrm{Var}_{P}\!\left(
\mathbb{E}_{R}[Y \mid D,P]
\right)
\right].
$$

Substituting this expression into the first equation gives the full three-component decomposition:
$$
\mathrm{Var}(Y)
=
\mathrm{Var}_{D}\!\left(
\mathbb{E}_{P,R}[Y \mid D]
\right)
+
\mathbb{E}_{D}\!\left[
\mathrm{Var}_{P}\!\left(
\mathbb{E}_{R}[Y \mid D,P]
\right)
\right]
+
\mathbb{E}_{D,P}\!\left[
\mathrm{Var}_{R}(Y \mid D,P)
\right].
$$

The three terms capture stable item difficulty, systematic prompt sensitivity, and run-to-run variability, respectively.
We assume that run-to-run variability is negligible under approximately deterministic decoding. Under this assumption,
$$
\mathbb{E}_{R}[Y \mid D,P] \approx Y(D,P)
\quad \text{and} \quad
\mathbb{E}_{D,P}\!\left[\mathrm{Var}_{R}(Y \mid D,P)\right] \approx 0.
$$
so the three-component expression reduces to the two-component decomposition used in Section \ref{sec:methodology}:
$$
\mathrm{Var}(Y)
\approx
\mathrm{Var}_{D}\!\left(
\mathbb{E}_{P}[Y \mid D]
\right)
+
\mathbb{E}_{D}\!\left[
\mathrm{Var}_{P}(Y \mid D)
\right].
$$

Because we use only one run per item and perturbation, any residual run-to-run variability, particularly for API models, cannot be estimated separately and is included in the observed within-item component. 
\textit{We therefore interpret the reported prompt component as an upper bound on variance attributable purely to prompt perturbations.} 

\textbf{Relationship to ANOVA:} 
We show that the variance decomposition used in the main text is equivalent to a random-effects ANOVA with perturbations nested within data items.
Let $Y_{d,p} \in \{0,1\}$ denote model correctness on data item $d$ under perturbation $p$. The evaluation protocol applies multiple perturbations to each data item, yielding a nested experimental design in which perturbations are nested within data items. Consider the random-effects model
\begin{equation*}
Y_{d,p} = \mu + \alpha_d + \beta_{p(d)},
\end{equation*}
where $\mu$ is the grand mean, $\alpha_d$ captures intrinsic item difficulty for data item $d$, and $\beta_{p(d)}$ captures sensitivity to perturbations applied to $d$. We treat both effects as random with zero mean. 
Under deterministic decoding (temperature 0), the residual term in Eq.(1) vanishes, with stochastic generation, it is absorbed into $\beta_{p(d)}$, consistent with the upper-bound interpretation in Section \ref{sec:methodology}.
Under this model, the total variance decomposes as
\begin{equation*}
\mathrm{Var}(Y) = \mathrm{Var}(\alpha_d) + \mathrm{Var}(\beta_{p(d)}).
\end{equation*}
These variance components correspond exactly to the variances terms of the methodology in the main text:
\begin{align*}
\mathrm{Var}(\alpha_d) &= \mathrm{Var}_D\!\left(\mathbb{E}_P[Y \mid D]\right), \\
\mathrm{Var}(\beta_{p(d)}) &= \mathbb{E}_D\!\left[\mathrm{Var}_P(Y \mid D)\right].
\end{align*}

Therefore, the proposed decomposition is equivalent to a variance-components ANOVA with perturbations nested within data items. The main text presents this decomposition at the population level via the law of total variance, whereas classical ANOVA estimates the same variance components from finite samples using sums of squares or likelihood-based methods.

\section{Inference and Post-processing Details}
\label{app:prompt}

For multiple-choice inference, we used a fixed system instruction that allows the model to perform free-form reasoning before emitting a final answer in a standardized XML-style tag. 
The user prompt remained the default template according to each task implementation in \texttt{lm-eval-harness}. 
Generation was run with a maximum sequence length of 32,768 tokens. We then applied a custom filter in \texttt{harness-eval} to parse the generated output and extract only the option enclosed in the \texttt{<answer>} tag. This ensured that evaluation was based solely on the model's explicitly selected answer, while allowing the preceding reasoning trace to vary in length.

\begin{tcolorbox}[
  colback=gray!5,
  colframe=black,
  title={Inference prompt for multiple-choice answer extraction},
  sharp corners,
  boxrule=0.5pt,
  fontupper=\footnotesize
]

\textbf{\underline{SYSTEM INSTRUCTION}}

\begin{verbatim}
You will be given a multiple-choice question. Before responding,
analyze the question and answer options. You can do any reasoning;
however, you MUST put your answer inside the answer tag:
<answer>A/B/C/D/</answer>. Do not include any other text on that line.
\end{verbatim}

\vspace{0.5em}
\label{app-inference-prompt}

\end{tcolorbox}

\section{Generalization Across Model Families}
\label{sec:extra-models}
To examine whether the main trends observed for Llama and Qwen extend to other model families, we conduct a targeted evaluation on three additional models. Rather than running the full \bb{} suite, we evaluate Gemma 3-4B \citep{team2025gemma}, Mistral-8B-Instruct \citep{liu2026ministral}, and Phi-4-mini-Instruct \citep{abdin2024phi} on a representative subset of 12 perturbations across three benchmarks (MMLU, TruthfulQA, and GPQA), resulting in 108 additional experiments. We report the results in Table~\ref{tab:main_results_new_models} in the same format as Table~\ref{tab:main_results}, stratified by benchmark and perturbation group.
The additional models broadly reproduce the main trends observed for the Llama and the Qwen families. Word-level perturbations reduce performance across all three model families, while prompt padding is particularly damaging for Mistral and Gemma. Prompt augmentation remains more model-dependent, producing a moderate drop for Mistral, little effect for Phi-4-mini, and a small average improvement for Gemma. These results suggest that the main patterns identified by \bb{} are not specific to the Llama and Qwen families, although the magnitude of sensitivity varies across models.

\begin{table*}
  \resizebox{1\textwidth}{!}{%
\begin{tabular}{l|c>{\columncolor[gray]{0.95}}c|c>{\columncolor[gray]{0.95}}c|c>{\columncolor[gray]{0.95}}c|c>{\columncolor[gray]{0.95}}c}
\toprule
\textbf{Model} & \multicolumn{2}{c}{\textbf{\begin{tabular}[c]{@{}c@{}}Word-level\\ Perturbations\end{tabular}}} & \multicolumn{2}{c}{\textbf{\begin{tabular}[c]{@{}c@{}}Prompt\\ Padding\end{tabular}}} & \multicolumn{2}{c}{\textbf{\begin{tabular}[c]{@{}c@{}}Prompt\\ Augmentation\end{tabular}}} & \multicolumn{2}{c}{\textbf{Average}} \\
               & \multicolumn{1}{c}{$\Delta Drop$ ($\downarrow$)} & \multicolumn{1}{c}{$Range$} 
               & \multicolumn{1}{c}{$\Delta Drop$ ($\downarrow$)} & \multicolumn{1}{l}{$Range$} 
               &  \multicolumn{1}{c}{$\Delta Drop$ ($\downarrow$)} & \multicolumn{1}{c}{ $Range$}
               & \multicolumn{1}{c}{$\Delta Drop$ ($\downarrow$)} & \multicolumn{1}{c}{ $Range$}
\\ 
\midrule
\multicolumn{9}{c}{\textit{\textit{Commercial Frontier Models}}} \\

Claude Opus 4.5 &  6.22 & 6.55 & 4.70 & 6.20 & 0.98 & 3.88 & 3.97 & 5.54 \\
GPT-5 &  12.24 & 12.67 & 0.06 & 0.65 & 2.97 & 2.87 & 5.09 & 5.40  \\

\midrule
\multicolumn{9}{c}{\textit{\textit{Open-Weight Models}}} \\

Qwen-32b & 7.52 & 7.30 & 6.05 & 1.71 & 1.64 & 1.05 & 5.07 & 3.35  \\
Qwen-8b &  7.06 & 6.32 & 11.87 & 11.55 & 0.96 & 1.45 & 6.63 & 6.44  \\
Qwen-4b &  7.97 & 8.23 & 10.63 & 10.15 & 1.39 & 1.64 & 6.66 & 6.67   \\
Llama3.3-70b & 7.61 & 8.32 & 10.17 & 8.62 & 2.68 & 3.34 & 6.82 & 6.76   \\
Llama3.1-70b & 6.49 & 6.47 & 11.98 & 5.87 & 2.94 & 4.55 & 7.14 & 5.63   \\
Llama3.1-8b & 5.01 & 4.18 & 15.19 & 17.74 & 5.20 & 8.94 & 8.47 & 10.29   \\

Mistral-8b & 2.79 & 3.16 & 17.74 & 18.23 & 4.79 & 6.00 & 8.44 & 9.13  \\
Phi-4-mini & 5.97 & 5.52 & 7.97 & 10.54 & 0.32 & 0.15 & 4.75 & 6.40 \\
Gemma3-4b & 5.46 & 5.29 & 16.60 & 17.00 & -0.71 & 6.29 & 7.12 & 9.53   \\
 \bottomrule
\end{tabular}
}
\caption{Mean $\Delta Drop$ and $Range$ (\%) per model and perturbation category across the six MCQ benchmarks used for \bb{}: \textit{MMLU, TruthfulQA and GPQA}.}
  \label{tab:main_results_new_models}
\end{table*}

\section{Uncertainty Analysis}
\label{sec:uncertainty}

To quantify the uncertainty associated with aggregating results across benchmarks, we compute confidence intervals for every cell in Table~\ref{tab:main_results}. Specifically, we perform a bootstrap analysis by resampling the six benchmarks with replacement over 10,000 iterations and taking the 2.5th and 97.5th percentiles of the resulting distribution. The resulting 95\% bootstrap confidence intervals capture benchmark-level uncertainty, that is, how much the reported $\Delta$Drop and Range vary across different resamples of the benchmark suite.

Table~\ref{tab:main_results_ci} reproduces Table~\ref{tab:main_results} with the corresponding confidence intervals.
The confidence intervals support the main trends reported in the main text of the paper. Word-level perturbations consistently produce performance drops across models, while prompt padding has particularly large effects on open-weight models. 
The intervals also support our characterization of paraphrasing as comparatively benign for most models, with several effects remaining close to zero.
Overall, the confidence intervals provide uncertainty estimates around the reported effects without changing the qualitative conclusions.

\section{Disaggregated Results}
\label{sec:results}

\subsection{Model performance on \bb}
Tables \ref{tab:res_word_manipulation}, \ref{tab:res_padding}, \ref{tab:res_augmentation}, and \ref{tab:res_paraphrasing} present a detailed breakdown of average model performance across all benchmarks, stratified by perturbation for each of the four perturbation groups. 
This disaggregated view isolates the effect of individual perturbations on model performance, enabling a fine-grained analysis at the perturbation level. 
Across these results, models exhibit broadly consistent trends, lending support to the validity of the perturbation groupings used in our evaluation.

\begin{table*}[h]
\centering
  \resizebox{1\textwidth}{!}{%
\begin{tabular}{l|c|c>{\columncolor[gray]{0.95}}c|c>{\columncolor[gray]{0.95}}c|c>{\columncolor[gray]{0.95}}c|c>{\columncolor[gray]{0.95}}c|c>{\columncolor[gray]{0.95}}c|c>{\columncolor[gray]{0.95}}c|ccc}
\toprule
\textbf{Model} & \textbf{Baseline} & \multicolumn{2}{c}{\textbf{Typos}} & \multicolumn{2}{c}{\textbf{\begin{tabular}[c]{@{}c@{}}Drop\\ Stop-words\end{tabular}}} & \multicolumn{2}{c}{\textbf{\begin{tabular}[c]{@{}c@{}}Punctuation\\ Spaces\end{tabular}}} & \multicolumn{2}{c}{\textbf{\begin{tabular}[c]{@{}c@{}}Sequence\\ Spaces\end{tabular}}} & \multicolumn{2}{c}{\textbf{\begin{tabular}[c]{@{}c@{}}Word\\ Merge\end{tabular}}}  &  \multicolumn{2}{c}{\textbf{\begin{tabular}[c]{@{}c@{}}Word\\ Split\end{tabular}}} &  \multicolumn{2}{c}{\textbf{}} \\

& & \multicolumn{1}{c}{Acc ($\uparrow$)} & \multicolumn{1}{c}{Drop ($\downarrow$)}
& \multicolumn{1}{c}{Acc ($\uparrow$)} & \multicolumn{1}{l}{Drop ($\downarrow$)} &
\multicolumn{1}{c}{Acc ($\uparrow$)} & \multicolumn{1}{c}{Drop ($\downarrow$)}
& \multicolumn{1}{c}{Acc ($\uparrow$)} & \multicolumn{1}{c}{Drop ($\downarrow$)}
& \multicolumn{1}{c}{Acc ($\uparrow$)} & \multicolumn{1}{c}{Drop ($\downarrow$)} 
& \multicolumn{1}{c}{Acc ($\uparrow$)} & \multicolumn{1}{c}{Drop ($\downarrow$)}  
& \multicolumn{1}{c}{$\Delta Drop$} & \multicolumn{1}{c}{$Range$}             
\\ 
\midrule
               
\multicolumn{16}{c}{\textit{\textit{Commercial Frontier Models}}} \\

Claude Opus 4.5 & 78.64 & 74.36 & 4.28 & 74.08 & 4.56 & 78.45 & 0.19 & 72.15 & 6.49 & 76.72 & 1.92 & 76.74 & 1.90 & 6.49 & 6.49 \\
GPT-5 & 73.97 & 64.65 & 9.32 & 65.63 & 8.34 & 70.29 & 3.68 & 69.47 & 4.50 & 70.19 & 3.78 & 70.42 & 3.55 & 9.32 & 9.32 \\

\midrule
\multicolumn{16}{c}{\textit{\textit{Open-Weight Models}}} \\

Qwen-32b & & & & & & & & & & & & & & & \\
    \hspace{0.5cm} 0\_shot & 81.53 & 73.99 & 7.54 & 76.48 & 5.05 & 81.42 & 0.11 & 80.06 & 1.47 & 81.36 & 0.17 & 81.02 & 0.51 & 7.54 & 7.53 \\
    \hspace{0.5cm} few\_shot & 77.88 & 59.62 & 18.26 & 74.25 & 3.63 & 78.88 & -1.00 & 78.34 & -0.46 & 79.30 & -1.42 & 78.99 & -1.11 & 18.26 & 19.69 \\
    
Qwen-8b & & & & & & & & & & & & & & & \\
    \hspace{0.5cm} 0\_shot & 78.50 & 70.00 & 8.50 & 72.56 & 5.94 & 78.25 & 0.25 & 77.27 & 1.23 & 78.52 & -0.02 & 78.04 & 0.46 & 8.50 & 8.52 \\
    \hspace{0.5cm} few\_shot & 76.72 & 67.88 & 8.84 & 71.01 & 5.71 & 75.41 & 1.31 & 75.37 & 1.35 & 75.80 & 0.92 & 76.20 & 0.52 & 8.84 & 8.84 \\
    
Qwen-4b & & & & & & & & & & & & & & & \\
    \hspace{0.5cm} 0\_shot & 75.71 & 66.44 & 9.27 & 68.87 & 6.84 & 75.69 & 0.02 & 73.22 & 2.49 & 75.07 & 0.64 & 74.07 & 1.64 & 9.27 & 9.27 \\
    \hspace{0.5cm} few\_shot & 75.01 & 65.99 & 9.02 & 69.74 & 5.27 & 74.86 & 0.15 & 73.08 & 1.93 & 75.20 & -0.19 & 73.26 & 1.75 & 9.02 & 9.20 \\
    
Llama3.3-70b & & & & & & & & & & & & & & & \\
    \hspace{0.5cm} 0\_shot & 66.41 & 58.92 & 7.49 & 59.42 & 6.99 & 66.46 & -0.05 & 60.92 & 5.49 & 65.85 & 0.56 & 64.03 & 2.38 & 7.49 & 7.55 \\
    \hspace{0.5cm} few\_shot & 67.70 & 61.55 & 6.15 & 62.59 & 5.11 & 67.63 & 0.07 & 61.85 & 5.85 & 66.76 & 0.94 & 65.30 & 2.40 & 6.15 & 6.16 \\
Llama3.1-70b & & & & & & & & & & & & & & & \\
    \hspace{0.5cm} 0\_shot & 64.67 & 57.59 & 7.08 & 58.28 & 6.39 & 64.52 & 0.15 & 59.11 & 5.56 & 64.02 & 0.65 & 62.06 & 2.61 & 7.08 & 7.08 \\
    \hspace{0.5cm} few\_shot & 63.39 & 57.54 & 5.85 & 58.06 & 5.33 & 63.49 & -0.10 & 58.31 & 5.08 & 62.05 & 1.34 & 61.04 & 2.35 & 5.85 & 5.94 \\
    
Llama3.1-8b & & & & & & & & & & & & & & & \\
    \hspace{0.5cm} 0\_shot & 43.48 & 34.16 & 9.32 & 37.33 & 6.15 & 42.38 & 1.10 & 34.06 & 9.42 & 41.03 & 2.45 & 37.20 & 6.28 & 9.42 & 9.42 \\
    \hspace{0.5cm} few\_shot & 33.13 & 27.27 & 5.86 & 30.09 & 3.04 & 33.87 & -0.74 & 26.65 & 6.48 & 32.65 & 0.48 & 30.08 & 3.05 & 6.48 & 7.22 \\

\bottomrule
\end{tabular}
}
  \caption{Average performance (\%) across all used benchmarks on \texttt{Brittlebench}  for \textbf{Word Manipulation Perturbations}. Each number is the average accuracy score on the 6 benchmarks used (\textit{MMLU, ARC, TruthfulQA, MathQA, GPQA, LogiQA}) based on their \texttt{harness-eval} implementations for different few-shot settings.}
  \label{tab:res_word_manipulation}
\end{table*}

\begin{table*}[h]
\centering
  \resizebox{0.80\textwidth}{!}{%
\begin{tabular}{l|c|c>{\columncolor[gray]{0.95}}c|c>{\columncolor[gray]{0.95}}c|c>{\columncolor[gray]{0.95}}c|cc}
\toprule
\textbf{Model} & \textbf{Baseline} & \multicolumn{2}{c}{\textbf{Persona}} & \multicolumn{2}{c}{\textbf{Emotion Prompt}} & \multicolumn{2}{c}{\textbf{Augmented Paraphrasing}} &  \multicolumn{2}{c}{\textbf{}} \\

& & \multicolumn{1}{c}{Acc ($\uparrow$)} & \multicolumn{1}{c}{Drop ($\downarrow$)} 
& \multicolumn{1}{c}{Acc ($\uparrow$)} & \multicolumn{1}{c}{Drop ($\downarrow$)} 
& \multicolumn{1}{c}{Acc ($\uparrow$)} & \multicolumn{1}{c}{Drop ($\downarrow$)} 
& \multicolumn{1}{c}{$\Delta Drop$} & \multicolumn{1}{c}{$Range$}             
              
\\ 
\midrule
               
\multicolumn{10}{c}{\textit{\textit{Commercial Frontier Models}}} \\

Claude Opus 4.5 & 78.64 & 77.68 & 0.96 & 78.05 & 0.59 & 77.08 & 1.56 & 1.56 & 1.56 \\
GPT-5 & 73.97 & 69.96 & 4.01 & 69.90 & 4.07 & 69.22 & 4.75 & 4.75 & 4.75 \\

\midrule
\multicolumn{10}{c}{\textit{\textit{Open-Weight Models}}} \\
Qwen-32b & & & & & & & & &   \\
    \hspace{0.5cm} 0\_shot & 81.53 & 81.18 & 0.35 & 80.70 & 0.83 & 81.21 & 0.32 & 0.83 & 0.83 \\
    \hspace{0.5cm} few\_shot & 77.88 & 80.36 & -2.48 & 79.35 & -1.47 & 74.74 & 3.14 & 3.14 & 5.62 \\
Qwen-8b & & & & & & & & &   \\
    \hspace{0.5cm} 0\_shot & 78.50 & 78.45 & 0.05 & 78.49 & 0.01 & 77.55 & 0.95 & 0.95 & 0.95 \\
    \hspace{0.5cm} few\_shot & 76.72 & 76.33 & 0.39 & 77.60 & -0.88 & 74.59 & 2.13 & 2.13 & 3.01 \\
Qwen-4b & & & & & & & & &  \\
    \hspace{0.5cm} 0\_shot & 75.71 & 74.92 & 0.79 & 75.04 & 0.67 & 75.33 & 0.38 & 0.79 & 0.79 \\
    \hspace{0.5cm} few\_shot & 75.01 & 74.10 & 0.91 & 75.49 & -0.48 & 74.69 & 0.32 & 0.91 & 1.39 \\
Llama3.3-70b & & & & & & & & &   \\
    \hspace{0.5cm} 0\_shot & 66.41 & 65.37 & 1.04 & 65.04 & 1.37 & 64.53 & 1.88 & 1.88 & 1.89 \\
    \hspace{0.5cm} few\_shot & 67.70 & 67.22 & 0.48 & 66.48 & 1.22 & 66.50 & 1.20 & 1.22 & 1.22 \\
Llama3.1-70b & & & & & & & & &   \\
    \hspace{0.5cm} 0\_shot & 64.67 & 59.36 & 5.31 & 62.56 & 2.11 & 66.81 & -2.14 & 5.31 & 7.45 \\
    \hspace{0.5cm} few\_shot & 63.39 & 48.65 & 14.74 & 61.70 & 1.69 & 59.29 & 4.10 & 14.74 & 14.74 \\
Llama3.1-8b & & & & & & & & &   \\
    \hspace{0.5cm} 0\_shot & 43.48 & 28.70 & 14.78 & 40.95 & 2.53 & 45.29 & -1.81 & 14.78 & 16.59 \\
    \hspace{0.5cm} few\_shot & 33.13 & 16.16 & 16.97 & 30.16 & 2.97 & 33.52 & -0.39 & 16.97 & 17.36 \\

\bottomrule
\end{tabular}
}
  \caption{Average performance (\%) across all used benchmarks on \texttt{Brittlebench} for \textbf{Prompt Augmentation Perturbations}. Each number is the average accuracy score on the 6 benchmarks used (\textit{MMLU, ARC, TruthfulQA, MathQA, GPQA, LogiQA}) based on their \texttt{harness-eval} implementations for different few-shot settings.}
  \label{tab:res_augmentation}
\end{table*}

\begin{table*}[h]
\centering
  \resizebox{0.8\textwidth}{!}{%
\begin{tabular}{l|c|c>{\columncolor[gray]{0.95}}c|c>{\columncolor[gray]{0.95}}c|c>{\columncolor[gray]{0.95}}c|ccc}

\toprule
\textbf{Model} & \textbf{Baseline} & \multicolumn{2}{c}{\textbf{Quotes}} & \multicolumn{2}{c}{\textbf{Spaces}} & \multicolumn{2}{c}{\textbf{New lines}} &  \multicolumn{2}{c}{\textbf{}} \\

& & \multicolumn{1}{c}{Acc ($\uparrow$)} & \multicolumn{1}{c}{Drop ($\downarrow$)}
& \multicolumn{1}{c}{Acc ($\uparrow$)} & \multicolumn{1}{l}{Drop ($\downarrow$)} 
&  \multicolumn{1}{c}{Acc ($\uparrow$)} & \multicolumn{1}{c}{Drop ($\downarrow$)}
& \multicolumn{1}{c}{$\Delta Drop$} & \multicolumn{1}{c}{$Range$}             
\\ 
\midrule
               
\multicolumn{10}{c}{\textit{\textit{Commercial Frontier Models}}} \\

Claude Opus 4.5 & 78.64 & 76.77 & 1.87 & 80.10 & -1.46 & 79.64 & -1.00 & 1.87 & 3.32 \\
GPT-5  & 73.97 & 73.20 & 0.77 & 74.80 & -0.83 & 73.86 & 0.11 & 0.77 & 1.60 \\

\midrule
\multicolumn{10}{c}{\textit{\textit{Open-Weight Models}}} \\

Qwen-32b & & & & & & & & & &  \\
    \hspace{0.5cm} 0\_shot & 81.53 & 76.89 & 4.64 & 76.12 & 5.41 & 79.56 & 1.97 & 5.41 & 5.41 \\
    \hspace{0.5cm} few\_shot & 77.88 & 70.06 & 7.82 & 75.90 & 1.98 & 78.09 & -0.21 & 7.82 & 8.03 \\
Qwen-8b & & & & & & & & & &  \\
    \hspace{0.5cm} 0\_shot & 78.50 & 62.03 & 16.47 & 78.54 & -0.04 & 76.93 & 1.57 & 16.47 & 16.51 \\
    \hspace{0.5cm} few\_shot & 76.72 & 49.38 & 27.34 & 73.46 & 3.26 & 69.72 & 7.00 & 27.34 & 27.34 \\
Qwen-4b & & & & & & & & & &  \\
    \hspace{0.5cm} 0\_shot & 75.71 & 68.54 & 7.17 & 71.83 & 3.88 & 76.14 & -0.43 & 7.17 & 7.60 \\
    \hspace{0.5cm} few\_shot & 75.01 & 68.04 & 6.97 & 72.18 & 2.83 & 75.44 & -0.43 & 6.97 & 7.40 \\
    
Llama3.3-70b & & & & & & & & & &  \\
    \hspace{0.5cm} 0\_shot & 66.41 & 59.76 & 6.65 & 63.25 & 3.16 & 66.74 & -0.33 & 6.65 & 6.98 \\
    \hspace{0.5cm} few\_shot & 67.70 & 63.53 & 4.17 & 66.81 & 0.89 & 67.06 & 0.64 & 4.17 & 4.17 \\
    
Llama3.1-70b & & & & & & & & & &  \\
    \hspace{0.5cm} 0\_shot & 64.67 & 54.16 & 10.51 & 58.36 & 6.31 & 56.57 & 8.10 & 10.51 & 10.51 \\
    \hspace{0.5cm} few\_shot & 63.39 & 47.18 & 16.21 & 54.60 & 8.79 & 53.79 & 9.60 & 16.21 & 16.21 \\
Llama3.1-8b & & & & & & & & & &  \\
    \hspace{0.5cm} 0\_shot & 43.48 & 21.87 & 21.61 & 39.92 & 3.56 & 40.19 & 3.29 & 21.61 & 21.61 \\
    \hspace{0.5cm} few\_shot & 33.13 & 10.67 & 22.46 & 21.00 & 12.13 & 12.81 & 20.32 & 22.46 & 22.46 \\

\bottomrule
\end{tabular}
}
  \caption{Average performance (\%) across all used benchmarks on \texttt{Brittlebench}  for \textbf{Prompt Padding Perturbations}. Each number is the average accuracy score on the 6 benchmarks used (\textit{MMLU, ARC, TruthfulQA, MathQA, GPQA, LogiQA}) based on their \texttt{harness-eval} implementations for different few-shot settings.}
  \label{tab:res_padding}
\end{table*}

\begin{table*}[h]
\centering
  \resizebox{0.85\textwidth}{!}{%
\begin{tabular}{l|c|c>{\columncolor[gray]{0.95}}c|c>{\columncolor[gray]{0.95}}c|c>{\columncolor[gray]{0.95}}c|ccc}

\toprule
\textbf{Model} & \textbf{Baseline} & \multicolumn{2}{c}{\textbf{Lexical Paraphrasing}} &\multicolumn{2}{c}{\textbf{Syntactical Paraphrasing}} & \multicolumn{2}{c}{\textbf{Rule-free Paraphrasing}} &  \multicolumn{2}{c}{\textbf{}} \\

& & \multicolumn{1}{c}{Acc ($\uparrow$)} & \multicolumn{1}{c}{Drop ($\downarrow$)}
& \multicolumn{1}{c}{Acc ($\uparrow$)} & \multicolumn{1}{c}{Drop ($\downarrow$)} 
&  \multicolumn{1}{c}{Acc ($\uparrow$)} & \multicolumn{1}{c}{Drop ($\downarrow$)}
& \multicolumn{1}{c}{$\Delta Drop$} & \multicolumn{1}{c}{$Range$}             
\\ 
\midrule
               
\multicolumn{10}{c}{\textit{\textit{Commercial Frontier Models}}} \\

Claude Opus 4.5 & 78.64 & 78.56 & 0.08 & 78.26 & 0.38 & 78.71 & -0.07 & 0.38 & 0.44 \\
GPT-5 & 73.97 & 69.14 & 4.83 & 68.31 & 5.66 & 69.42 & 4.55 & 5.66 & 5.66 \\

\midrule
\multicolumn{10}{c}{\textit{\textit{Open-Weight Models}}} \\

Qwen-32b & & & & & & & & &\\
    \hspace{0.5cm} 0\_shot & 81.53 & 81.08 & 0.45 & 81.04 & 0.49 & 81.30 & 0.23 & 0.49 & 0.48 \\
    \hspace{0.5cm} few\_shot & 77.88 & 76.58 & 1.30 & 77.26 & 0.62 & 76.40 & 1.48 & 1.48 & 1.48 \\
Qwen-8b & & & & & & & & &\\
    \hspace{0.5cm} 0\_shot & 78.50 & 78.30 & 0.20 & 77.90 & 0.60 & 78.08 & 0.42 & 0.60 & 0.60 \\
    \hspace{0.5cm} few\_shot & 76.72 & 73.95 & 2.77 & 73.57 & 3.15 & 73.60 & 3.12 & 3.15 & 3.14 \\
Qwen-4b & & & & & & & & &\\
    \hspace{0.5cm} 0\_shot & 75.71 & 75.15 & 0.56 & 74.53 & 1.18 & 75.67 & 0.04 & 1.18 & 1.18 \\
    \hspace{0.5cm} few\_shot & 75.01 & 74.41 & 0.60 & 74.24 & 0.77 & 74.39 & 0.62 & 0.77 & 0.76 \\
    
Llama3.3-70b & & & & & & & & &\\
    \hspace{0.5cm} 0\_shot & 66.41 & 65.58 & 0.83 & 65.53 & 0.88 & 65.83 & 0.58 & 0.88 & 0.88 \\
    \hspace{0.5cm} few\_shot & 67.70 & 66.32 & 1.38 & 66.15 & 1.55 & 66.56 & 1.14 & 1.55 & 1.55 \\
Llama3.1-70b & & & & & & & & &\\
    \hspace{0.5cm} 0\_shot & 64.67 & 66.59 & -1.92 & 67.04 & -2.37 & 66.81 & -2.14 & -0.01 & 2.37 \\
    \hspace{0.5cm} few\_shot & 63.39 & 59.63 & 3.76 & 59.14 & 4.25 & 60.11 & 3.28 & 4.25 & 4.25 \\
Llama3.1-8b & & & & & & & & & \\
    \hspace{0.5cm} 0\_shot & 43.48 & 46.21 & -2.73 & 45.26 & -1.78 & 46.71 & -3.23 & 0.00 & 3.23 \\
    \hspace{0.5cm} few\_shot & 33.13 & 34.64 & -1.51 & 33.97 & -0.84 & 35.00 & -1.87 & 0.00 & 1.87 \\

\bottomrule
\end{tabular}
}
  \caption{Average performance (\%) across all used benchmarks on \texttt{Brittlebench}  for \textbf{Paraphrasing Perturbations}. Each number is the average accuracy score on the 6 benchmarks used (\textit{MMLU, ARC, TruthfulQA, MathQA, GPQA, LogiQA}) based on their \texttt{harness-eval} implementations for different few-shot settings.}
  \label{tab:res_paraphrasing}
\end{table*}

\begin{table*}[ht]
\centering
  \resizebox{0.70\textwidth}{!}{
\begin{tabular}{l c | l c}
\toprule
\textbf{Zero-shot Perturbation} & \textbf{Rank Corr.} & \textbf{Few-shot Perturbation} & \textbf{Rank Corr.} \\
\midrule
punct\_spaces      & 0.9102 & word\_split      & 0.9673 \\
rule\_free\_par    & 0.8857 & punct\_spaces    & 0.9592 \\
word\_split        & 0.8776 & typos            & 0.9510 \\
lexical\_par       & 0.8694 & word\_merge      & 0.9429 \\
spaces             & 0.8694 & seq\_spaces      & 0.9347 \\
syntactic\_par     & 0.8449 & drop\_stopwords  & 0.9347 \\
seq\_spaces        & 0.8286 & lexical\_par     & 0.9347 \\
word\_merge        & 0.8286 & emotion\_prompt  & 0.9265 \\
persona            & 0.8204 & syntactic\_par   & 0.9265 \\
typos              & 0.8204 & rule\_free\_par  & 0.9184 \\
drop\_stopwords    & 0.8041 & augmented\_par   & 0.9102 \\
augmented\_par     & 0.8041 & persona          & 0.9020 \\
emotion\_prompt    & 0.8041 & spaces           & 0.8122 \\
quotes             & 0.6735 & new\_lines       & 0.7388 \\
new\_lines         & 0.6327 & quotes           & 0.7388 \\
\bottomrule
\end{tabular}
}
\caption{Mean Spearman rank correlation by perturbation for zero-shot and few-shot settings. Averaged across tasks.}
\label{tab:rank_correlation}
\end{table*}


\section{Open-Generation Tasks}
\label{sec:gen_tasks}

We extend our analysis to generation-based, leaderboard-relevant benchmarks, specifically HumanEval \citep{chen2021codex} (code generation) and GSM8K \citep{cobbe2021gsm8k} (math reasoning), using their standard implementations in \texttt{lm-evaluation-harness}. 

\begin{table*}
  \resizebox{1\textwidth}{!}{%
\begin{tabular}{l|c>{\columncolor[gray]{0.95}}c|c>{\columncolor[gray]{0.95}}c|c>{\columncolor[gray]{0.95}}c|c>{\columncolor[gray]{0.95}}c}
\toprule
\textbf{Model} & \multicolumn{2}{c}{\textbf{\begin{tabular}[c]{@{}c@{}}Word-level\\ Perturbations\end{tabular}}} & \multicolumn{2}{c}{\textbf{\begin{tabular}[c]{@{}c@{}}Prompt\\ Padding\end{tabular}}} & \multicolumn{2}{c}{\textbf{\begin{tabular}[c]{@{}c@{}}Prompt\\ Augmentation\end{tabular}}} & \multicolumn{2}{c}{\textbf{Paraphrasing}} \\
               & \multicolumn{1}{c}{$\Delta Drop$ ($\downarrow$)} & \multicolumn{1}{c}{$Range$} 
               & \multicolumn{1}{c}{$\Delta Drop$ ($\downarrow$)} & \multicolumn{1}{l}{$Range$} 
               &  \multicolumn{1}{c}{$\Delta Drop$ ($\downarrow$)} & \multicolumn{1}{c}{ $Range$}
               & \multicolumn{1}{c}{$\Delta Drop$ ($\downarrow$)} & \multicolumn{1}{c}{ $Range$}
\\ 
\midrule
\multicolumn{9}{c}{\textit{\textit{Commercial Frontier Models}}} \\

Claude Opus 4.5 &  6.49 [4.22, 10.12] & 6.49 [4.22, 10.15] & 1.87 [0.15, 4.77] & 3.32  [1.04, 7.10]	 & 1.56 [0.32, 4.50] & 1.56 [0.50, 4.69] & 0.38 [0.00, 1.49] & 0.44 [0.28, 2.00] \\
GPT-5 & 9.32 [4.08, 17.22] & 9.32 [4.08, 17.22] & 0.77 [0.00, 10.91] & 1.60 [0.49, 12.56]	 & 4.75 [0.58, 11.56] & 4.75 [0.64, 11.56] & 5.66 [0.69, 12.35] & 5.66 [1.24, 12.35] \\

\midrule
\multicolumn{9}{c}{\textit{\textit{Open-Weight Models}}} \\

Qwen-32b & 7.54 [3.19, 15.60] & 7.54 [3.50, 15.80] & 5.41 [4.28, 7.11] & 5.41  [4.28, 7.11] & 0.83 [0.54, 1.46] & 0.83 [0.58, 1.74] & 0.49 [0.17, 1.18] & 0.49 [0.19, 1.18]\\
Qwen-8b & 8.50 [3.47, 17.54] & 8.52 [3.90, 17.55] & 16.47 [10.42, 24.38] & 16.51 [10.42, 25.77] & 0.95  [0.15, 2.54]	& 0.95 [0.26, 3.24] & 0.60 [0.06, 1.27] & 0.60 [0.17, 1.34] \\
Qwen-4b &  9.27 [5.00, 17.47] & 9.27 [5.00, 17.47] & 7.17 [4.54, 10.82] & 7.60[5.11, 11.10] & 0.79 [0.62, 1.82] & 0.79 [0.62, 2.21] & 1.18 & 1.18  \\
Llama3.3-70b &  7.49 [5.40, 15.18] & 7.55 [5.46, 15.24] & 6.65  [2.55, 13.09] & 6.98 [4.29, 13.28] & 1.88  [0.88, 4.05] & 1.89 [0.90, 4.33] & 0.88 [0.18, 1.95] & 0.88 [0.22, 1.95] \\
Llama3.1-70b &  7.08 [4.16, 15.47] & 7.08 [4.30, 15.47] & 10.51 [5.61, 16.41] & 10.51 [5.70, 16.41]	 & 5.31 [1.83, 10.16] & 7.45 [3.73, 12.00]	& -0.01 [0.00, 0.98] & 2.37 [0.45, 7.19] \\
Llama3.1-8b &  9.42 [5.01, 17.17] & 9.42 [5.01, 17.17] & 21.61 [15.08, 28.48] & 21.61 [15.20, 28.59] & 14.78 [9.05, 20.29] & 16.59 [11.16, 21.26] & 0.00 [0.00, 1.06] & 3.23 [0.65, 7.08] \\

\bottomrule
\end{tabular}
}
\caption{Mean $\Delta Drop$ and $Range$ (\%) per model and perturbation category across the six MCQ benchmarks used for \bb{}: \textit{MMLU, ARC, TruthfulQA, MathQA, GPQA, and LogiQA}, along with 95\% bootstrap confidence intervals computed by resampling benchmarks.}
  \label{tab:main_results_ci}
\end{table*}

\begin{table*}[ht]
\centering
\resizebox{0.55\textwidth}{!}{
\begin{tabular}{llrrr}
\toprule
\textbf{Model} & \textbf{Benchmark} & \textbf{Perturbation} & \textbf{Performance} & \textbf{Drop} \\
\midrule
\multirow{5}{*}{Qwen-32B} & \multirow{5}{*}{HumanEval}
  & \textbf{baseline}         & \textbf{37.20} & \textbf{--} \\
& & word-level                & 0.92           & 36.28 \\
& & padding                   & 30.69          & 6.51 \\
& & prompt-augm               & 17.75          & 19.45 \\
& & \textbf{all perturbations} & \textbf{16.45} & \textbf{20.75} \\
\midrule
\multirow{5}{*}{Qwen-32B} & \multirow{5}{*}{GSM8K}
  & \textbf{baseline}         & \textbf{65.05} & \textbf{--} \\
& & word-level                & 50.17          & 14.88 \\
& & padding                   & 20.47          & 44.58 \\
& & prompt-augm               & 53.72          & 11.33 \\
& & \textbf{all perturbations} & \textbf{41.45} & \textbf{23.60} \\
\midrule
\multirow{5}{*}{Llama-3.3-70B} & \multirow{5}{*}{HumanEval}
  & \textbf{baseline}         & \textbf{47.56} & \textbf{--} \\
& & word-level                & 1.83           & 45.73 \\
& & padding                   & 36.79          & 10.77 \\
& & prompt-augm               & 40.19          & 7.37 \\
& & \textbf{all perturbations} & \textbf{26.27} & \textbf{21.29} \\
\midrule
\multirow{5}{*}{Llama-3.3-70B} & \multirow{5}{*}{GSM8K}
  & \textbf{baseline}         & \textbf{53.75} & \textbf{--} \\
& & word-level                & 50.53          & 3.22 \\
& & padding                   & 49.23          & 4.52 \\
& & prompt-augm               & 65.58          & -11.83 \\
& & \textbf{all perturbations} & \textbf{55.11} & \textbf{-1.36} \\
\bottomrule
\end{tabular}
}
\caption{Impact of semantics-preserving prompt perturbations on generation-based benchmarks (HumanEval and GSM8K) for Qwen-32B and Llama-3.3-70B in the zero-shot setting. We report baseline performance and signed performance changes relative to the baseline under different perturbation categories. Negative drops indicate improvements over the baseline.}
\label{tab:gen_results}
\end{table*}

The results in Table \ref{tab:gen_results} show that prompt brittleness clearly extends beyond MCQ benchmarks to generation-based evaluations, with distinct and model-dependent failure modes. 
On HumanEval, performance is highly sensitive to surface-form perturbations, with word-level changes causing near-complete collapse due to models mechanically propagating perturbations (mimicking) into the generated code, thereby breaking code executability rather than reasoning itself. For example, simple word splits in the prompt can lead to invalid outputs such as \colorbox{blue_back}{\texttt{return numbe r - i nt(numbe r)}}, which fail unit tests despite preserving the intended logic. Prompt augmentation and formatting perturbations further contribute to degradation, though to a lesser extent.
On GSM8K, the effects are more nuanced: Qwen-32B exhibits substantial brittleness, especially to padding, despite no obvious structural corruption, indicating instability in reasoning, whereas Llama-3.3-70B remains comparatively robust and even benefits from some perturbations (e.g., augmentation).

Overall, these results reveal both syntactic brittleness (code generation) and reasoning brittleness (math), and importantly, the magnitude and direction of these effects vary across models, directly demonstrating that perturbations can meaningfully shift relative performance even in realistic, leaderboard-relevant settings.

\clearpage

\section{Chain-of-thought and Reasoning Analysis}
\label{sec:app_reasoning}


We examine two key questions regarding the relationship between reasoning and brittleness: first, we ask \textit{\textbf{whether} reasoning can mitigate brittleness}. Cognitive scientists have found that deliberative reasoning and explanation allow people to discover and use generalizable principles as opposed to brittle heuristics \citep{williams2010role}. To examine whether reasoning can provide similar mitigation in LLMs, we leverage chain-of-thought (CoT) prompting, which is known to lead to more deliberative responses \citep{wei2022chain}.

Second, we ask \textit{\textbf{when} reasoning mitigates brittleness}. Studies of distraction show that reasoning primarily benefits performance when it is not itself compromised through distraction---for instance, semantic distractions lead to sharper drops in task performance than perceptual distractions in cognitive tasks \citep{kucwaj2022various}. We therefore examined whether the quality of reasoning can explain when it successfully mitigates brittleness, in particular, whether reasoning boosted performance when it was of high quality, and lowered performance when it was compromised by perturbations.


\subsection{Does Reasoning Mitigate or Cause Brittleness?}
Across all our analyses of the impact of reasoning, we examined a frontier model, Claude Opus 4.5. We first examined the effect of chain-of-thought (CoT) on its accuracy across our benchmarks. We found that CoT improved overall accuracy, though with variance across benchmarks (see Figure \ref{fig:CoTperformance}). Across 83,870 questions, CoT prompting significantly outperformed Non-CoT (McNemar's $\chi^2 = 787.71$, $p < 0.001$), with CoT answering correctly on 3,413 questions (4.1\%) where Non-CoT failed, compared to only 1,454 questions (1.7\%) where Non-CoT succeeded and CoT failed. The lack of improvement we observed for MathQA and TruthfulQA could either be due to ceiling effects, or related to past research showing heterogeneity in returns to CoT \citep{liu2024mind}. Note that across all analyses in this section, we used regular expressions to extract reasoning model's answers from their output strings; where extraction failed for a given question, both the CoT and non-CoT computations exclude that question.

\begin{figure}[]
    \centering
    \includegraphics[width=1\linewidth]{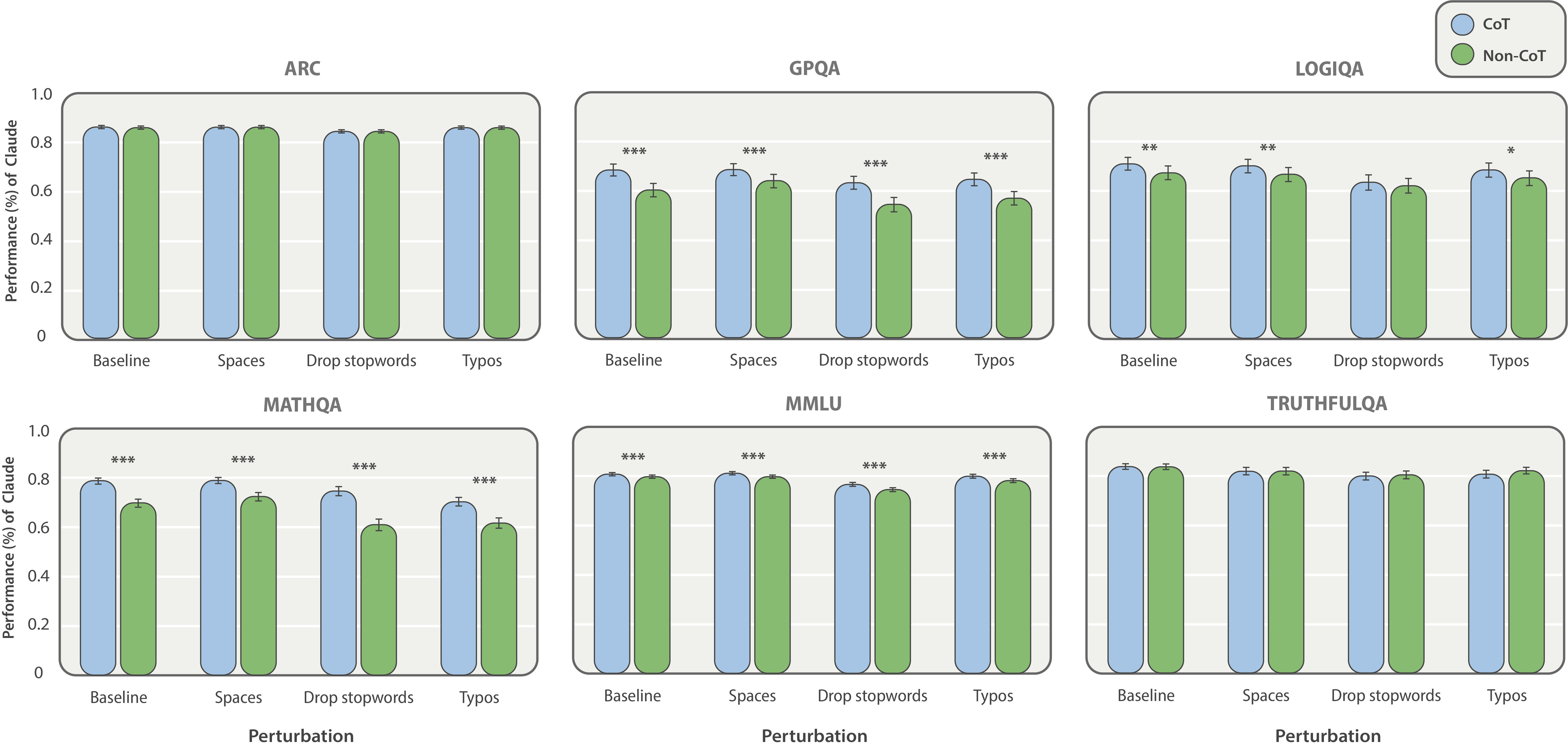} 
    \caption{CoT improves robustness across most, but not all benchmarks, for Claude Opus 4.5.}
    \label{fig:CoTperformance}
\end{figure}

These results broadly show that CoT boosts performance across benchmarks. However, they do not address whether CoT mitigates robustness---for that, we examine whether the drop in performance following perturbations is lower with CoT, and whether this occurs across benchmarks and perturbation types. We therefore conducted a mixed-effects regression analysis, where we regressed the accuracy of each inference on whether it was generated via CoT, whether it came from a perturbed prompt, and the interaction of these factors as fixed effects, and benchmark and perturbation type as random effects. If CoT mitigates brittleness, the interaction effect should be significant, and this prediction is borne out in our data ($\beta = .007, p = .006$, see Table \ref{tab:app_CoT_mixed_effects}).

These results show that, beyond merely boosting overall accuracy, CoT specifically mitigates robustness. Note that the practical size of this effect is quite small (the drop in performance from baseline to perturbed conditions was $2.38\%$ for CoT (from $93.02\%$ to $90.64\%$) and $2.79\%$ for Non-CoT (from $92.00\%$ to $89.21\%$). So while these CoT results suggest that reasoning may protect against brittleness, further research needs to explore many important questions \textit{which kinds} of reasoning architectures or interventions are most effective, and \textit{whether} models with routers choose to deliberate when they stand to benefit from it \citep{oktar2022deciding}.

\begin{table}[htbp]
\centering
\resizebox{0.55\textwidth}{!}{%
\begin{tabular}{lcccc}
\toprule
& Estimate & SE & $z$ & $p$ \\
\midrule
\multicolumn{5}{l}{\textit{Fixed Effects}} \\
\quad Intercept (Non-CoT, Baseline) & 0.880 & 0.038 & 23.01 & $<.001^{***}$ \\
\quad CoT (vs Non-CoT) & 0.012 & 0.002 & 5.12 & $<.001^{***}$\\
\quad Perturbed (vs Baseline) & -0.039 & 0.044 & -0.89 & .373 \\
\quad CoT $\times$ Perturbed & 0.008 & 0.003 & 2.81 & .005$^{**}$ \\
\\
\multicolumn{5}{l}{\textit{Random Effects (Variance / SD)}} \\
\quad Benchmark (Intercept) & \multicolumn{2}{c}{$<.001$} & \multicolumn{2}{c}{$<.001$} \\
\quad Perturbation (Intercept) & \multicolumn{2}{c}{0.0088} & \multicolumn{2}{c}{0.0936} \\
\quad Residual & \multicolumn{2}{c}{0.0828} & \multicolumn{2}{c}{0.2878} \\
\\
\multicolumn{5}{l}{\textit{Model Information}} \\
\quad Observations & \multicolumn{4}{c}{247,612} \\
\quad Groups (Benchmark) & \multicolumn{4}{c}{6} \\
\quad Groups (Condition) & \multicolumn{4}{c}{4} \\
\quad Log-Likelihood & \multicolumn{4}{c}{-43049.52} \\
\bottomrule
\end{tabular}
}
\caption{CoT weakly mitigates against brittleness in perturbations. The regression uses accuracy (0/1) as the dependent variable. The key predictor is the CoT $\times$ Perturbed interaction: a positive coefficient indicates that perturbation hurts CoT \textit{less} than Non-CoT, meaning CoT is more robust. Model includes random intercepts for benchmark and condition. Note that the significance of direct effects is difficult to interpret in the presence of interactions.
$^*p<.05$, $^{**}p<.01$, $^{***}p<.001$.}
\label{tab:app_CoT_mixed_effects}
\end{table}

\subsection{When Does Reasoning Mitigate or Cause Brittleness?}
One explanation for why CoT improves robustness is that reasoning can allow the model to extract, represent, and process the perturbed input with greater fidelity. This explanation requires that the reasoning produced be `good' itself, however---if perturbations also corrupt reasoning traces, we should not expect CoT to boost performance. To examine this possibility, we rated the quality of reasoning in each trace using GPT-5 Thinking as LLM-as-judge. This judge outputs a `reasoning quality score' (0-100) that evaluates features such as the coherence and completeness of each reasoning trace through a rubric (see Section \ref{app_reasoning_judge} for details on the judge, including the verbatim prompt). 

For drops in reasoning quality to explain drops in accuracy, perturbations need to affect reasoning itself. Our analyses show that such drops are common across benchmarks and perturbations (see Figure \ref{fig:CoTreasoningquality}). The effects that we observe within most benchmarks also translate to significant overall decrease in reasoning quality with perturbations ($\beta = -2.61, p<.001$; see Table \ref{tab:reasoning_quality_regression} for details). 

There remains an important question: do these reasoning judgments track meaningful signal about the quality of the reasoning? That is, it could be that perturbations lower the quality of reasoning, but that model accuracy is largely insensitive to these shifts. Moreover, establishing a link between the quality of reasoning and accuracy would provide face validity to the meaningfulness of the reasoning quality evaluation of our judge model. Our analyses show that there is a significant and substantial relationship between reasoning quality scores and accuracy, with the correct answer being associated with higher quality reasoning across benchmarks ($\beta = 0.011, p < .001$). Taken together with the previous result that perturbations lower reasoning quality, these analyses support a simple mechanism: reasoning can mitigate the effects of perturbations primarily when it is not itself compromised by the perturbation. 

\begin{figure}[]
    \centering
    \includegraphics[width=1\linewidth]{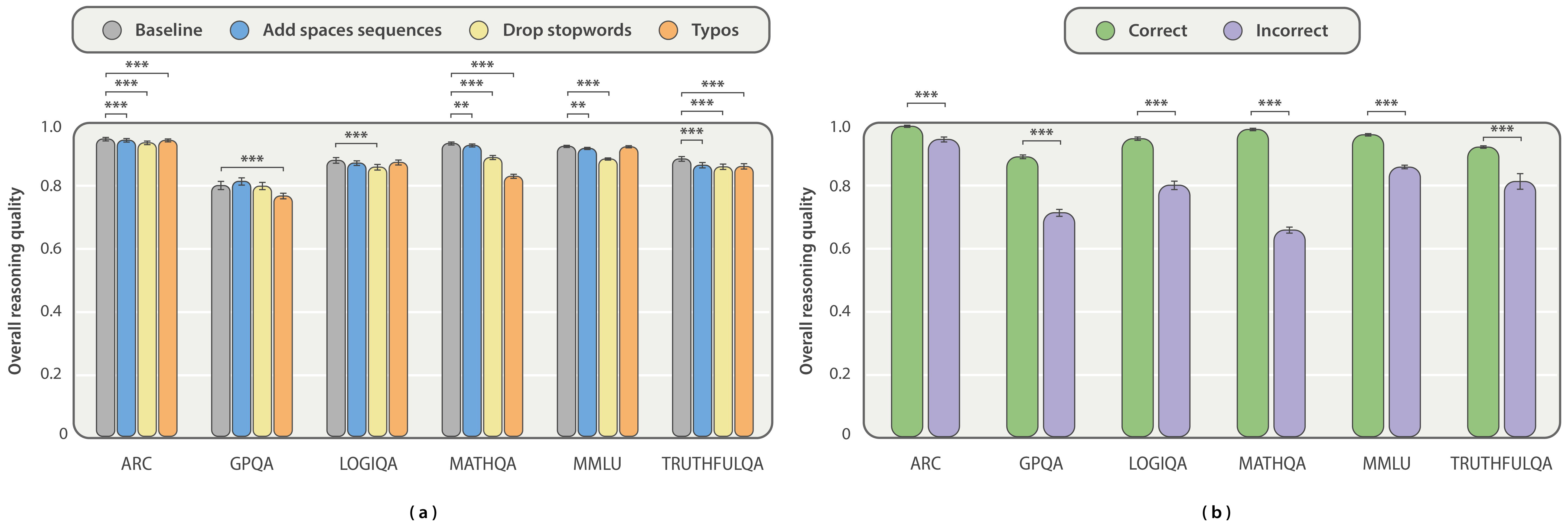} 
    \caption{\textbf{(a):} Perturbations cause drops to reasoning quality, though with variation across benchmarks in the size of the drop. \textbf{(b):} Correct answers are systematically linked with high reasoning scores across all benchmarks.}
    \label{fig:CoTperformance}
\end{figure}


\begin{table}[htbp]
\centering
\resizebox{0.65\textwidth}{!}{%
\begin{tabular}{lcccc}
\toprule
& Estimate & SE & $z$ & $p$ \\
\midrule
\multicolumn{5}{l}{\textit{Fixed Effects}} \\
\quad Intercept (Baseline) & 89.21 & 2.14 & 41.69 & $<.001^{***}$ \\
\quad Perturbed (vs Baseline) & -2.15 & 0.07 & -30.99 & $<.001^{***}$ \\
\\
\multicolumn{5}{l}{\textit{Random Effects (Variance)}} \\
\quad Benchmark & \multicolumn{4}{c}{27.43} \\
\quad Residual & \multicolumn{4}{c}{121.21} \\
\\
\multicolumn{5}{l}{\textit{Model Information}} \\
\quad Observations & \multicolumn{4}{c}{123,329} \\
\quad Log-Likelihood & \multicolumn{4}{c}{-470854.02} \\
\bottomrule
\end{tabular}
}
\caption{Reasoning quality is significantly lower for perturbations than baseline. The regression uses overall reasoning quality (0--100 scale) as the DV. 
Model includes random intercepts for benchmark.
$^*p<.05$, $^{**}p<.01$, $^{***}p<.001$.}
\label{tab:reasoning_quality_regression}
\end{table}

\clearpage

\subsection{Judge Implementation Details}
\label{app_reasoning_judge}

Below, we provide details on how the reasoning judge was implemented. We first provide the prompt, which describes the rubric and instructions used by the LLM to evaluate reasoning traces. 
As the judge outputs multiple dimensions, we examined the relationship between these to decide whether to analyze them independently, and found them all to be highly correlated (all $r \geq .9$); we therefore decided to use the overall quality score (which had the highest pairwise correlations with the other scores; all $r \geq .94$) as the reasoning score.

\begin{tcolorbox}[
  colback=gray!5,
  colframe=black,
  title={Judge prompt for CoT Reasoning traces evaluation},
  sharp corners,
  boxrule=0.5pt,
  fontupper=\footnotesize
]
You are an expert judge evaluating the quality of a reasoning trace produced by an LLM on a multiple-choice question. \\

\textbf{\underline{CONTEXT}}

\textbf{QUESTION/PROBLEM:}

\texttt{\{question\}}

\vspace{0.5em}
\textbf{ANSWER OPTIONS:}

\texttt{\{options\}}

\vspace{0.5em}
\textbf{CORRECT ANSWER:}

\texttt{\{correct\_answer\}}

\vspace{0.5em}
\textbf{MODEL'S REASONING TRACE (final answer stripped to avoid bias):}

\texttt{\{prediction\}}

\vspace{1em}
\textbf{\underline{EVALUATION CRITERIA}}

Evaluate the reasoning trace on the following dimensions (0--100 scale each).

IMPORTANT: Focus ONLY on the quality of the reasoning process itself, NOT on whether the model got the correct answer.

\vspace{0.5em}
\textbf{1. LOGICAL COHERENCE (0--100)}
\begin{itemize}
    \item Does the reasoning follow a logical, step-by-step structure?
    \item Are there any logical fallacies, contradictions, or non-sequiturs?
    \item Does each step follow from the previous one?
\end{itemize}
\textit{LOW (0--30):} Incoherent, contradictory, or illogical reasoning

\textit{MEDIUM (31--70):} Some logical flow but with gaps or minor inconsistencies

\textit{HIGH (71--100):} Clear, logically sound progression of ideas

\vspace{0.5em}
\textbf{2. STEP COMPLETENESS (0--100)}
\begin{itemize}
    \item Are all necessary intermediate steps shown?
    \item Is any critical reasoning step missing?
    \item Is the chain of thought complete from problem to answer?
\end{itemize}
\textit{LOW (0--30):} Major steps missing, jumps to conclusion

\textit{MEDIUM (31--70):} Most steps present but some gaps

\textit{HIGH (71--100):} All relevant steps clearly articulated

\vspace{0.5em}
\textbf{3. CLARITY (0--100)}
\begin{itemize}
    \item Is the reasoning easy to follow?
    \item Are variables/concepts clearly defined?
    \item Is the writing clear and well-organized?
\end{itemize}
\textit{LOW (0--30):} Confusing, poorly structured, hard to follow

\textit{MEDIUM (31--70):} Understandable but could be clearer

\textit{HIGH (71--100):} Crystal clear, well-organized explanation

\vspace{0.5em}

\end{tcolorbox}

\begin{tcolorbox}[
  colback=gray!5,
  colframe=black,
  title={Judge prompt for CoT Reasoning traces evaluation},
  sharp corners,
  boxrule=0.5pt,
  fontupper=\footnotesize
]

\textbf{4. ANSWER JUSTIFICATION (0--100)}
\begin{itemize}
    \item Does the reasoning properly justify the chosen answer?
    \item Is there a clear connection between the analysis and the final answer?
    \item Are alternative answers properly ruled out (if applicable)?
\end{itemize}
\textit{LOW (0--30):} Answer appears disconnected from reasoning

\textit{MEDIUM (31--70):} Some justification but weak connection

\textit{HIGH (71--100):} Strong, clear justification for the answer

\vspace{1em}
\textbf{\underline{INSTRUCTIONS}}

Provide your evaluation in EXACTLY this format:

\begin{verbatim}
LOGICAL_COHERENCE: [0-100]
STEP_COMPLETENESS: [0-100]
CLARITY: [0-100]
ANSWER_JUSTIFICATION: [0-100]
OVERALL_QUALITY: [0-100]
BRIEF_EXPLANATION: [2-3 sentences explaining your overall assessment]
\end{verbatim}
\label{app-reasoning-judge-prompt}

\end{tcolorbox}

\section{Limitations}
\label{sec:limitations}

Our variance decomposition is defined over observed binary correctness rather than latent model confidence or answer quality. 
This choice makes the metric comparable across open-weight and API-only models, since not all models expose token probabilities or calibrated likelihoods. 
However, binary correctness does not capture changes in confidence, reasoning quality, or cases where a perturbation changes one incorrect answer into another incorrect answer.
As a result, our brittleness scores measure correctness-level instability and may underestimate broader forms of model sensitivity. 
Brittleness scores also depend on the chosen perturbation taxonomy: a more aggressive set would mechanically increase $V_{prompt}$, so our scores are comparable across models within our taxonomy rather than taxonomy-invariant properties of a model.
Since we use a single inference run per prompt variant, inference-level stochasticity cannot be fully separated from perturbation-induced variation. Finally, our perturbation taxonomy is broad but not exhaustive, and real-world prompts may contain more complex combinations of noise, ambiguity, and conversational context than those studied here. Estimating this instability also requires repeated inference over semantically equivalent variants, introducing an evaluation-efficiency trade-off that we discuss further in Appendix~\ref{app:impact}.

\section{Impact Statement}
\label{app:impact}

As has become well known, generative models (such as LLMs) generally cost more to evaluate than other kinds of AI models, as running large model inference is usually required. Since, the \bb{} framework creates and evaluates semantically-equivalent variants of existing datasets, it has the potential to explode the cost of running inference, as more perturbations, more perturbation types\footnote{Moreover, for each kind of perturbation, the space of possible variants is likely to be large (we have provided some samples from the space of possible perturbations for our perturbation types, but surely more are possible).}, more benchmarks, or more models are investigated. This limitation is not specific to our work; it also crops up for adversarial robustness \citep{silva-najafirad-2020-Opportunities, carlini-etal-2019-evaluating}, automated jail-breaking \citep{li-etal-2025-efficient} and other related ideas \citep{kiela-etal-2021-dynabench, papakipos-etal-2022-augly}. To combat this potential combinatorial explosion, we recommend focused, follow-up effort to improve inference efficiency for robustness evaluation. One could draw inspiration from works such as \citep{rodriguez-etal-2021-evaluation, varshney-etal-2022-ildae, vivek-etal-2024-anchor, gupta-etal-2025-improving}, which aim to improve sample efficiency or shrink benchmark size through filtering, subset correlation, dataset instance analysis and other means, while maintaining discriminative power and informativeness.

\end{document}